\documentclass[10pt,twocolumn,letterpaper]{article}

\usepackage{iccv}
\usepackage{times}
\usepackage{epsfig}
\usepackage{graphicx}
\usepackage{amsmath}
\usepackage{amssymb}

\usepackage{multirow}
\usepackage{bm}
\usepackage{colortbl}
\usepackage{soul}
\definecolor{Gray}{gray}{0.9}
\sethlcolor{Gray}
\usepackage{enumitem}
\usepackage{subcaption}

\usepackage{algpseudocode}
\usepackage{algorithm}
\usepackage[pagebackref=true,breaklinks=true,letterpaper=true,colorlinks,bookmarks=false]{hyperref}

\iccvfinalcopy 

\def\iccvPaperID{**} 

\ificcvfinal\fi

\begin{document}

\title{DiffRate : Differentiable Compression Rate for Efficient Vision Transformers}

\author{Mengzhao Chen$^{1,2,3\dagger}$, Wenqi Shao$^{3*}$, Peng Xu$^{3,4}$, Mingbao Lin$^{5}$, Kaipeng Zhang$^3$, Fei Chao$^{1,2}$, \\Rongrong Ji$^{1,2*}$, Yu Qiao$^{3}$, Ping Luo$^{3,4}$ \\
$^1$Key Laboratory of Multimedia Trusted Perception and Efficient Computing,
\\Ministry of Education of China, Xiamen University\\
$^2$Institute of Artificial Intelligence, Xiamen University \\
$^3$OpenGVLab, Shanghai AI Laboratory  $^4$The University of Hong Kong  
$^5$Tencent Holdings Ltd 
}

\maketitle
\renewcommand{\thefootnote}{\fnsymbol{footnote}}
{\let\thefootnote\relax\footnotetext{
\noindent \hspace{-5mm}$^*$Corresponding authors: Rongrong Ji (rrji@xmu.edu.cn), Wenqi Shao (shaowenqi@pjlab.org.cn) \\
$^\dagger$ This work was done during his internship at Shanghai AI Laboratory. }}
\ificcvfinal\thispagestyle{empty}\fi

\begin{abstract}
Token compression aims to speed up large-scale vision transformers (\eg ViTs) by pruning (dropping) or merging tokens. It is an important but challenging task.
Although recent advanced approaches achieved great success, they need to carefully handcraft a compression rate (\ie number of tokens to remove), which is tedious and leads to sub-optimal performance.
To tackle this problem, we propose \textbf{Diff}erentiable Compression \textbf{Rate} (DiffRate), a novel token compression method that has several appealing properties prior arts do not have. First, DiffRate enables propagating the loss function's gradient onto the compression ratio, which is considered as a non-differentiable hyperparameter in previous work. In this case, different layers can automatically learn different compression rates layer-wisely without extra overhead. 
Second, token pruning and merging can be naturally performed simultaneously in DiffRate, while they were isolated in previous works. 
Third, extensive experiments demonstrate that DiffRate achieves state-of-the-art performance. For example, by applying the learned layer-wise compression rates to an off-the-shelf ViT-H (MAE) model, we achieve a 40\% FLOPs reduction and a 1.5$\times$ throughput improvement, with a minor accuracy drop of 0.16\% on ImageNet without fine-tuning, even outperforming previous methods with fine-tuning. Codes and models are available at \url{https://github.com/OpenGVLab/DiffRate}.

\end{abstract}

\section{Introduction}

\begin{figure}[!ht]
    \centering
    \includegraphics[width=0.95\linewidth]{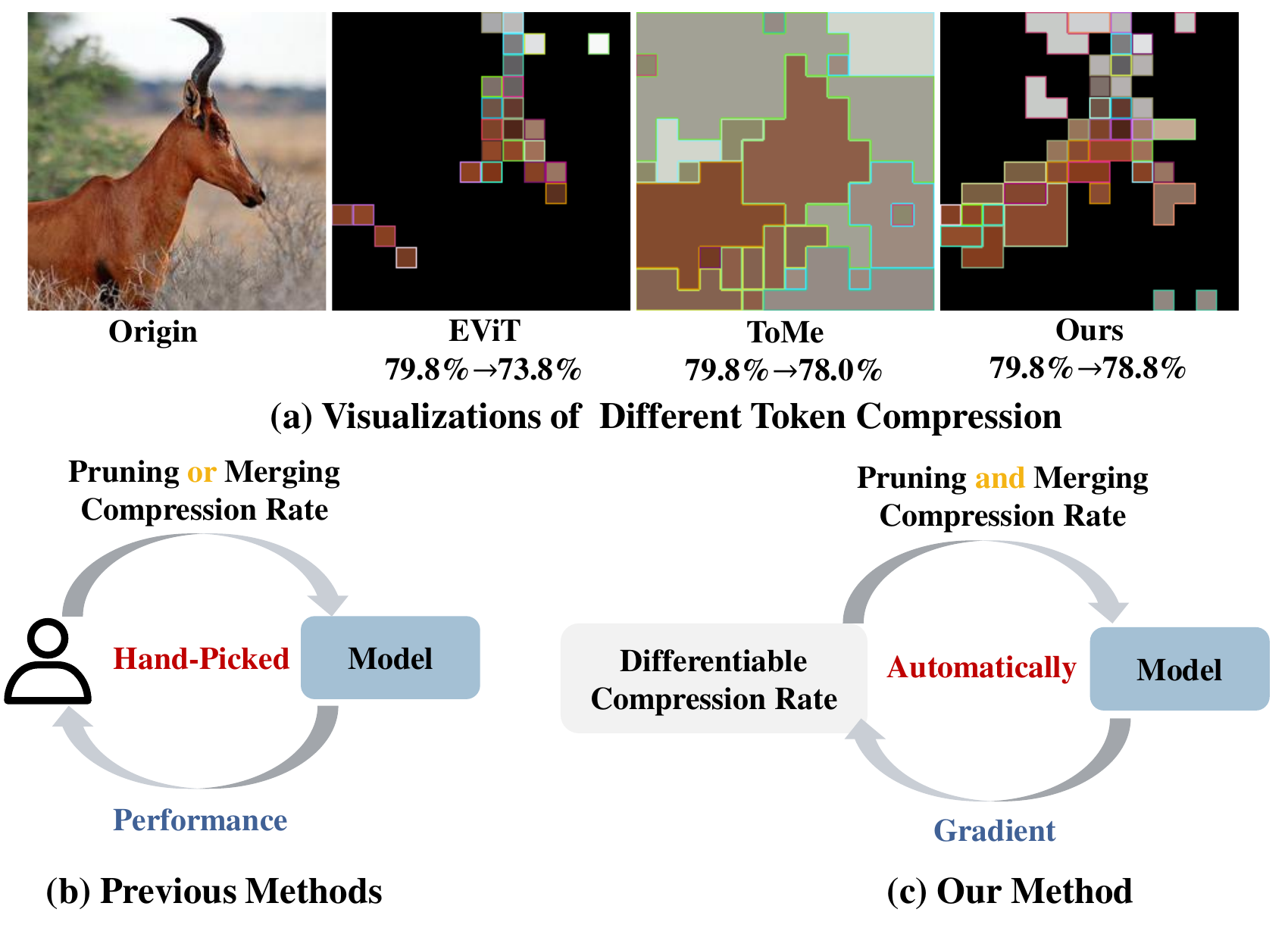}
    \vspace{-1em}
    \caption{\textbf{Comparison of different token compression methods} including token pruning EViT~\cite{liang2022not}, token merging ToMe~\cite{bolya2022tome} and our method. Pruned tokens are represented by black and non-border, while merged tokens are represented by patches with the same inner and border color. (\textbf{a}) shows that our method achieves better top-$1$ accuracy on ImageNet with FLOPs of $2.3$G when compressing pretrained Deit-S~\cite{touvron2021training} without fine-tuning. (\textbf{b}) and (\textbf{c}) show that previous methods typically focus on either pruning or merging tokens using hand-picked compression rate with the guidance of performance. But our method leverages both approaches simultaneously to achieve more effective compression using the differentiable compression rate with gradient optimization.}
    \label{fig:introduction}
    \vspace{-2em}
\end{figure}

Vision Transformer (ViT)~\cite{dosovitskiy2020image} has rapidly developed and achieved state-of-the-art performance in various vision tasks such as image classification\cite{liu2021swin}, object detection \cite{zhu2022deformable}, and semantic segmentation  \cite{xie2021segformer,han2022survey}. Due to the flexibility in handling various input formats, ViT has also been widely applied to self-supervised learning~\cite{he2022masked} and other modalities~\cite{feichtenhofer2022masked,xu2022masked}. Despite the remarkable success, ViTs suffer from the intensive computational complexity that increases quadratically with the token length in the self-attention layer, presenting a challenge for practical applications. Therefore, it is crucial to improve the efficiency of ViTs in order to make them more widespread.

In pursuit of efficient ViTs, various network compression techniques such as weight pruning~\cite{yang2021nvit,wangvtc}, quantization~\cite{liu2021post,yuan2021ptq4vit}, distillation~\cite{jia2021efficient,wu2022tinyvit} and so on have been investigated. Among them, token compression\cite{rao2021dynamicvit,bolya2022tome,liang2022not} has emerged as a promising approach to reduce redundancy in ViT for several appealing properties. First, token compression can be applied without any modifications to the network structure by exploiting the input-agnostic nature of transformers. Second, token compression is orthogonal to previous compression methods, making it a complementary approach to existing techniques \cite{wangvtc,chavan2022slimming}.
Existing token compression approaches typically include token pruning \cite{rao2021dynamicvit,liang2022not,kong2022spvit} and token merging \cite{bolya2022tome}. As shown in Fig.\,\ref{fig:introduction}(a), token pruning preserves informative tokens by measuring the importance of tokens with a defined metric. It can easily identify irrelevant background tokens with low importance. On the other hand, token merging compresses tokens by merging them with a large semantic similarity, which can not only discard some background tokens but also merge less informative foreground tokens. 
However, both pruning and merging handcraft a compression rate (\ie the ratio of removed tokens and total tokens)  for each transformer layer as shown in Fig.\,\ref{fig:introduction}(b), which has two drawbacks.
First, since model complexity metrics such as FLOPs are related to compression rates in each layer, it is tedious for practitioners to set layer-wise compression rates in order to meet the complexity constraints while retaining the performance of ViTs as much as possible. Second, informative foreground tokens are prone to being discarded with a hand-picked compression rate, which results in performance degradation. As shown in Fig.\,\ref{fig:introduction}(a), when compressing tokens at fixed rates, token pruning such as EViT~\cite{liang2022not} removes most of the informative foreground tokens while token merging \cite{bolya2022tome} also merges many important foreground tokens into a single token, leading to a sudden drop of top-$1$ accuracy on ImageNet.

%

To tackle the above issues, this work proposes a unified token compression framework called Differentiable Compression Rate (DiffRate), where both pruning and merging compression rates are determined in a differentiable manner. 
To achieve this goal, we propose a novel method, namely Differentiable Discrete Proxy (DDP) module. In DDP, a token sorting procedure is first performed to identify important tokens with a token importance metric. Then, a re-parameterization trick enables us to optimally select top-$K$ important tokens with gradient back-propagation. In this way, all input images would have the top-$K$ important tokens preserved, making it possible for parallel batch computation. Notably, the optimization process of DiffRate is highly efficient and can converge within $3$ epochs (\ie 2.7 GPU hours for ViT-B).
%

Thanks to the inclusion of differentiable compression rates, DiffRate can leverage the benefits of token pruning and merging by seamlessly integrating both techniques into a forward pass. This is possible because both token pruning and merging are capable of determining the optimal set of tokens to preserve.
As shown in Fig.\,\ref{fig:introduction}(a), 
DiffRate can prune most irrelevant background tokens and preserve detailed foreground information, leading to a good trade-off between efficiency and performance.
 With the learned compression rate, DiffRate achieves state-of-the-art performance in compressing various ViTs.
 For example, DiffRate can compress an off-the-shelf ViT-H model pre-trained by MAE \cite{he2022masked} with $40$\% FLOPs reduction and $50$\% throughput improvement with only $0.16$\% accuracy drop, outperforming previous methods that require tuning the network parameter.
 

Our contributions are summarized as follows:
\begin{itemize}
\item We develop a unified token compression framework, Differentiable Compression Rate (DiffRate), that includes both token pruning and merging, and formulate token compression as an optimization problem. 
\vspace{-0.3em}
\item DiffRate employs a Differentiable Discrete Proxy which consists of a token sorting procedure and a re-parameterization trick to determine the optimal compression rate under different computation cost constraints. To our knowledge, it is the first study to explore differentiable compression rate optimization in token compression.
\vspace{-0.3em}
\item Through extensive experiments, we demonstrate that DiffRate outperforms previous methods and achieves state-of-the-art performance on the off-the-shelf models. We hope that DiffRate can advance the field of token compression and improve the practical application of Vision Transformers (ViTs).  
\end{itemize}

\section{Related Work}

\textbf{Token Compression} Several recent studies have attempt compress redundancy token according token pruning~\cite{liang2022not,rao2021dynamicvit,fayyaz2022ats,yin2022vit,tang2022patch,li2022sait,kong2022spvit,xu2022evo,lin2022super,wangvtc} and token merging~\cite{bolya2022tome,zeng2022not,ryoo2021tokenlearner,marin2021token}. However, most of these methods focus on designing metrics to distinguish redundant tokens, while ignoring the token compression schedule in each block. Some methods, such as VTC-LFC~\cite{wangvtc} and ViT-Slim~\cite{chavan2022slimming} combine token prune with weight prune and determine the number of prune tokens using threshold-based approaches, which is also highly influenced by the hand-picked hyperparameters.
In contrast, our proposed method can learn the token compression schedule in a differentiable form. Furthermore, we consider both token merging and token pruning in the token compression process, resulting in a better trade-off between speed and accuracy.

\textbf{Differentiable Neural Architecture Search.} There are also many works~\cite{liu2018darts,cai2018proxylessnas,guo2020dmcp} attempted to search for neural architecture in a differentiable manner. For example, DARTS~\cite{liu2018darts} and ProxylessNAS~\cite{cai2018proxylessnas} learn the probabilities of each candidate operation and select the operation with the highest probability as the final architecture. DMCP~\cite{guo2020dmcp} is the approach most similar to proposed DiffRate, as it searches for the channel number of convolution in each layer. 
However, the differentiable method used in channel pruning cannot be directly applied to token compression.
Firstly, the definition of search space is different. The search space for channel pruning is the channel, while for token compression, it is the token. Channels in the same position represent the same feature, while tokens are position-agnostic.
Secondly, the output channel number in each layer is independent to other layers while the token must be pruned once it is discarded in previous layers.
%
As of now, no approach has been proposed for differentiable token compression rate.


\section{Differentiable Compression Rate}
In this section, we first briefly introduce a transformer block and existing compression approaches, and then present our Differentiable Compression Rate (DiffRate) to build a unified token compression technique.

\textbf{Transformer Block.} Token compression in ViTs operates in each transformer block. Given the input token of the $l$-th block $\mathbf{X}^l\in \mathbb{R}^{N\times D}$ where $N$ and $D$ are the token length and token size, respectively, the forward propagation of the transformer block is expressed as follows:
\begin{equation} \label{eq:trasnformer}
    \hat{\mathbf{X}}^l = \mathbf{X}^l + \mathrm{Attention}(\mathbf{X}^{l}),\\
    \mathbf{X}^{l+1} = \hat{\mathbf{X}}^{l} + \mathrm{MLP}(\hat{\mathbf{X}}^{l}),
\end{equation}
where $l\in [L]$ and $L$ is the network depth. Moreover, $\mathrm{Attention}$ and $\mathrm{MLP}$ represent the self-attention and the MLP modules in the transformer block, respectively. In Eqn.\,(\ref{eq:trasnformer}), $\hat{\mathbf{X}}^l$ is the output token of Attention.

\begin{figure}[!t]
    \centering
    \includegraphics[width=0.95\linewidth]{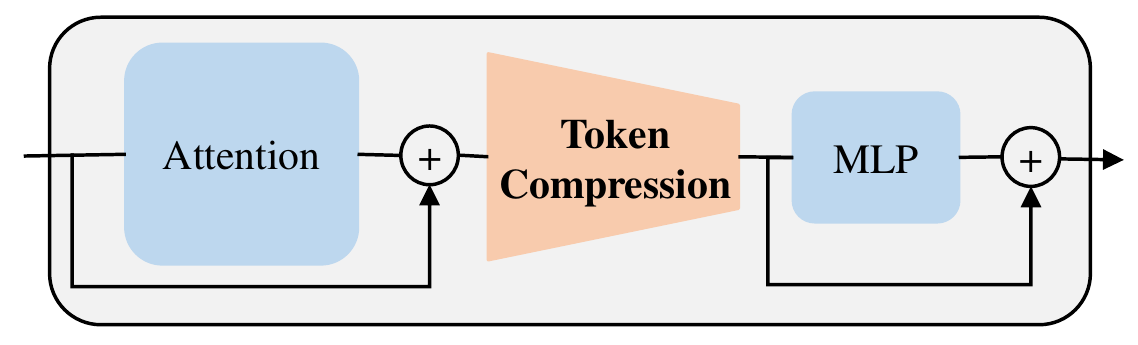}
    \vspace{-1em}
    \caption{\textbf{Token compression} in a transformer block. 
    }
    \label{fig:token_compression}
    \vspace{-1.5em}
\end{figure}

\textbf{Token Pruning and Merging.} As shown in Fig.\,\ref{fig:token_compression}, existing token compression methods usually remove redundant tokens from $\hat{\mathbf{X}}^l$ by token pruning or merging, as given by
\begin{equation}\label{eq:pruning-merging}
        \hat{\mathbf{X}}^{l}_p \leftarrow f_{p}(\hat{\mathbf{X}}^{l}, \alpha_{p}^{l})\,\,\mathrm{or} \,\,
    \hat{\mathbf{X}}^{l}_m \leftarrow f_{m}(\hat{\mathbf{X}}^{l}, \alpha_{m}^{l}),
\end{equation}
where $f_{p}, f_{m}$ are pruning and merging operations,$\alpha_{p}^{l}, \alpha_{m}^{l}$ are their compression rates, and $\hat{\mathbf{X}}^{l}_p\in \mathbb{R}^{N^l_p\times D}, \hat{\mathbf{X}}^{l}_m\in \mathbb{R}^{N^l_m\times D}$ are their outputs which are then fed into MLP in Eqn.\,(\ref{eq:trasnformer}). Hence, the pruning and merging compression rate for each block is defined as $\alpha_{p}^{l}=(N-N^l_p)/N$ and $ \alpha_{m}^{l}=(N-N^l_m)/N$, respectively.
For example, EViT~\cite{liang2022not} preserves the important tokens while fusing unimportant tokens between Attention and MLP under the guidance of an importance metric. ToMe~\cite{bolya2022tome} merges similar tokens of $\hat{\mathbf{X}}^{l}$ in both foreground and background. Note that DynamicViT~\cite{rao2021dynamicvit} prunes tokens after MLP, but we find that it also works well when it operates after Attention.  Although these approaches achieved great success, they need to carefully handcraft a compression rate block-wisely, which is tedious and leads to sub-optimal performance as shown in Fig.\,\ref{fig:introduction}(a).

\textbf{Unified Formulation of DiffRate.} To mitigate this problem, we propose Differentiable Compression Rate (DiffRate), a unified token compression method, to search compression rates optimally.
Given a pre-trained model $\mathbf{W}^{*}$, the objective of token compression is to minimize the classification loss $\mathcal{L}_{cls}$ on a training dataset $(\mathbf{X},\mathbf{Y})$ with target FLOPs $T$.
This can be formulated as an optimization problem as follows:
\begin{align}
\bm{\alpha}^*_{p},\bm{\alpha}_{m}^* =\arg\min_{\bm{\alpha}_{p},\bm{\alpha}_{m}} \mathcal{L}_{cls}(\mathbf{W}^{*}(\mathbf{X}),\mathbf{Y} | \bm{\alpha}_p,\bm{\alpha}_m), \label{eq:loss} \\
\text{s.t. } \mathcal{F}(\bm{\alpha}_{p},\bm{\alpha}_{m}) \leq T, 0 \leq \alpha_{p}^l,\alpha_{m}^l \leq 1,\label{eq:constraint} \\
\hat{\mathbf{X}}^{l} = f_{c}(\hat{\mathbf{X}}^{l},\alpha_p^l,\alpha_m^l), l\in[L] \label{eq:forward}
\end{align}
where $\bm{\alpha}_{p}=\{\alpha_{p}^{l}\}_{l=1}^L$ and $\bm{\alpha}_{m}=\{\alpha_{m}^{l}\}_{l=1}^L$ represent pruning and merging compression rates across all blocks, respectively. Moreover, $\mathcal{F}(\bm{\alpha}_{p},\bm{\alpha}_{m})$ denotes the corresponding FLOPs, which can be expressed as a differentiable way of the compression rates. Eqn.\,(\ref{eq:forward}) shows that DiffRate compresses $\hat{\mathbf{X}}^l$ with operation $f_c$ and compression rates $\alpha_p^l$ and $\alpha_m^l$ in each transformer block. Finally, $\bm{\alpha}^*_{p}$ and $\bm{\alpha}_{m}^*$ are obtained by differentiably learning in DiffRate.

With the unified formulation of token compression, DiffRate is capable enough to express various compression methods. In detail, when $f_c=f_p, \alpha_m^l=0$, DiffRate represents token pruning with differentiable pruning compression rate $\alpha_p^l$. when $f_c=f_m, \alpha_p^l=0$, DiffRate turn into differentiable token merging. In this work, we set $f_c=f_m\circ f_p$, meaning that tokens are first pruned and then merged. In this case, DiffRate seamlessly integrates token pruning and token merging through differentiable compression rates.

However, it is challenging to solve the optimization problem in Eqn.\,(\ref{eq:loss}-\ref{eq:forward}) with gradient-based methods for ViTs as compression rates are not differentiable. Directly learning $0$-$1$ masks for tokens as did in channel pruning~\cite{kang2020operation} is infeasible because each image may drop different numbers of tokens. This makes it hard to parallelize the computation. For example, DynamicViT~\cite{rao2021dynamicvit} and SPViT~\cite{kong2022spvit} maintain a mask vector for each input image, but they still need to manually design compression rates to ensure that all images preserve the same number of tokens. 
The following section introduces a novel technique for the differentiable search of compression rates.


\begin{figure*}[!htbp]
  \centering
  \begin{subfigure}[b]{0.45\linewidth}
    \includegraphics[width=\linewidth]{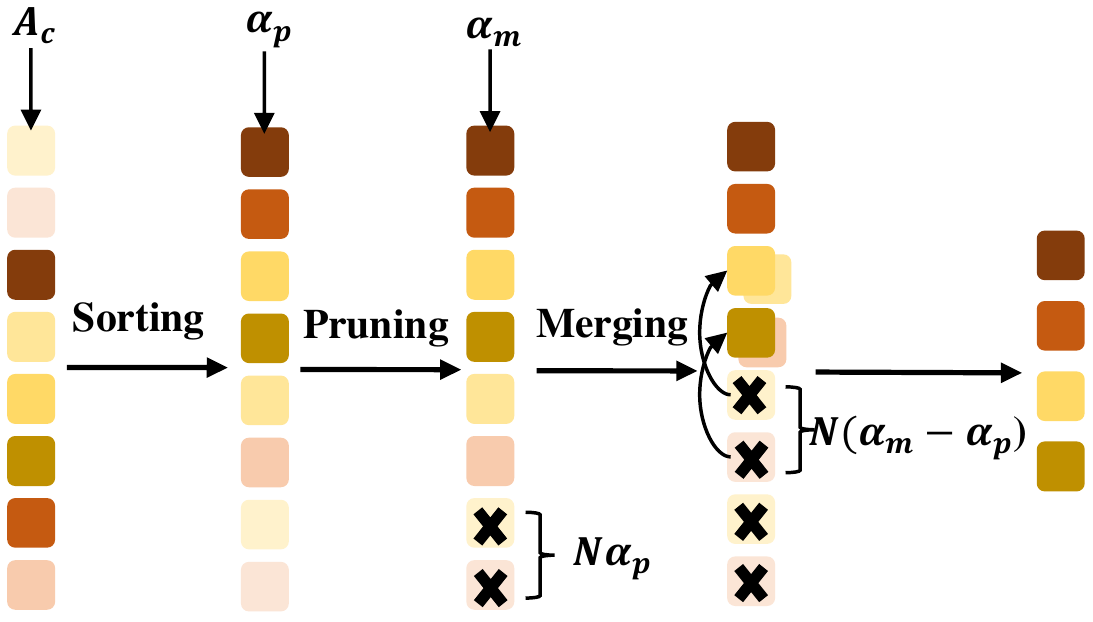}
    \caption{Token Sorting}
    \label{fig:token_sorting}
  \end{subfigure}
  \quad
  \begin{subfigure}[b]{0.45\linewidth}
    \includegraphics[width=\linewidth]{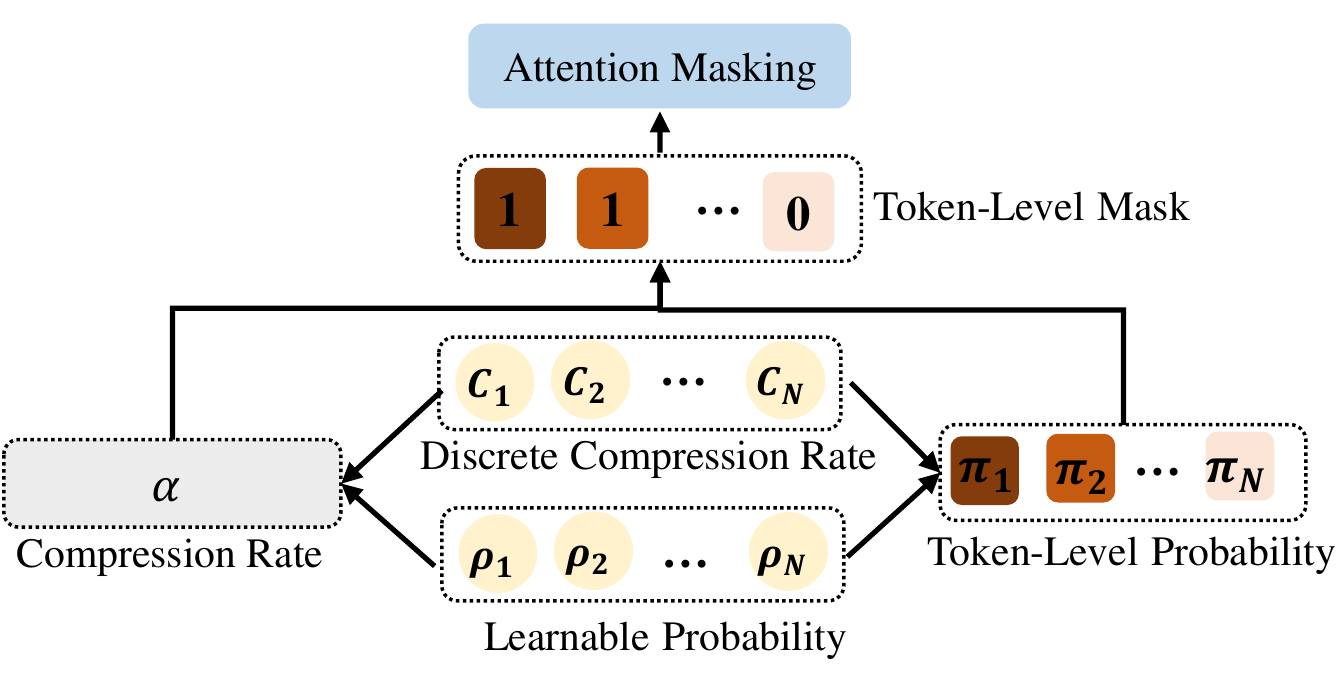}
    \caption{Compression Rate Re-parameterization}
    \label{fig:reparameteriza}
  \end{subfigure}
    \vspace{-0.5em}
  \caption{\textbf{Pipeline of Differentiable Discrete Proxy}. (a) Token Sorting: the input $N$ tokens are sorted based on the importance metric class attention $A_{c}$. With pruning rate $\alpha_p$ and merging rate $\alpha_m$, we first prune $N\alpha^p$ least important tokens, then merge $N(\alpha^m - \alpha^p)$  unimportant tokens with similar ones among the remaining tokens.  (b) Compression Rate Re-parameterization: the approach translate compression rate $\alpha$ into the combination with discrete rate $\mathbf{C}$ and learnable probability $\rho$. The top part is attention masking~\cite{rao2021dynamicvit}, which simulates token dropping by mask during training.}
  \label{fig:pipeline}
    \vspace{-1em}
\end{figure*}

\section{Differentiable Discrete Proxy}

To make compression rates differentiable, our main idea is to preserve the top-$K$ important tokens for all images, which can not only allow for parallel batch computation but also retain the performance of the original ViTs as much as possible. To achieve this, we introduce a novel method called the Differentiable Discrete Proxy (DDP), which comprises two critical components: a token sorting procedure to identify important tokens with a token importance metric and a re-parameterization trick to optimally select top-$K$ important tokens with gradient back-propagation. The overall pipeline of DDP is illustrated in  Fig.\,\ref{fig:pipeline}.

\subsection{Token Sorting}\label{sec:token-sort}
\textbf{Token Importance Metric.} To find top-$K$ importance tokens, we sort tokens by token importance metric, which has been well established in the literature. Here, we employ the class attention $\mathbf{A}_{c}\in\mathbb{R}^{1\times N}$ as the importance metric following EViT \cite{liang2022not}.
The interaction between class attention and image tokes can be written by:
\begin{equation}
    \mathbf{A}_{c} = \mathrm{Softmax}\big(\mathbf{q}_{c}\mathbf{K}^T/\sqrt{D}\big),\,\,\mathrm{and} \,\, \mathbf{X}_{c} = \mathbf{A}_{c} \mathbf{V},\label{eq:importance-metric}
\end{equation}
where $\mathbf{q}_{c}\in\mathbb{R}^{1\times D}$, $\mathbf{K}\in\mathbb{R}^{N\times D}$, $\mathbf{V}\in\mathbb{R}^{N\times D}$ and $\mathbf{X}_{c}\in\mathbb{R}^{1\times D}$ denote the query vector of class token, the key matrix, the value matrix, and the class token of the self-attention layer. 
From Eqn.\,(\ref{eq:importance-metric}), the class attention $\mathbf{A}_{c}$ measures how much each image token contributes to the class token. Higher class attention indicates a more significant influence of the corresponding image token on the final output, implying greater importance~\cite{chen2022coarse,liang2022not}.  
We also investigate other importance metrics in the ablation study as shown in Table\,\ref{tab:sorting_metric}.

\textbf{Pruning and Merging in DiffRate.} 
With the token importance established in Eqn.\,(\ref{eq:importance-metric}), it is natural to remove tokens with low importance, such as those representing semantically irrelevant backgrounds by following principles in token pruning~\cite{liang2022not,rao2021dynamicvit}. As shown in Fig.\,\ref{fig:token_sorting}, we prune $N\alpha_p$ unimportant tokens in the $l$-th transformer block.
Notely, we drop the superscript $l$ of compression rate to simplify the notation.
After that, we use cosine similarity to measure the similarity between $N(\alpha_m-\alpha_p)$ unimportant tokens and the remaining tokens. For similar token pairs, we generate a new token by directly average them.
Through the above sorting-pruning-merging pipeline, the number of tokens to be pruned and merged in each block is optimally determined with learnable compression rate in our DiffRate. Hence, DiffRate can seamlessly integrate token pruning and merging.

\subsection{Compression Rate Re-parameterization}\label{sec:reparam}
DDP uses a re-parameterization trick to make pruning and merging compression rates differentiable. We simplify the notation by using a single variable $\alpha$ to represent both compression rates.

%
\textbf{Re-parameterization with Discrete Rates.} 
%
In essential, making compression rate differentiable is to determine how many tokens should be discarded with optimality guarantee. To tackle this problem, we re-parameterize the compression rate as a learnable combination of multiple candidate compression rates.
Specifically, we introduce a discrete compression rate set, denoted as $\mathbf{C} = \{{C}_1, {C}_2, ..., {C}_N\}$, where $C_k=\frac{k-1}{N}$  represents the top $(k-1)$ least important tokens should be removed. By assigning learnable probabilities ${\rho}_k$ to each candidate compression rate ${C}_k$ with  $\sum_{k=1}^{N}\rho_k = 1$, the compression rate can be written as 
\begin{equation}
    \alpha = \sum_{k=1}^{N} {C}_{k} {\rho}_{k}.\label{eq:expected_keeping_ratio}
\end{equation} 
By using discrete candidate rates as the proxy, the optimization problem of learning compression rates can be translated into the problem of learning the probabilities ${\rho_k}$.

\textbf{Token-Level Mask.}
As shown in Fig.\,\ref{fig:reparameteriza}, with ${C}_k$ and ${\rho}_k$, the probability that the $k$-th important token is compressed can be calculated as 
\begin{equation}\label{eq:token-level-prob}
   \pi_1 =0, {\pi}_k = \rho_{N+2-k}+\cdots + \rho_{N-1}+ {\rho}_{N}, k\geq 2
\end{equation}
where $\mathbf{\pi}_1=0$ indicates that the most important token is always retained. From Eqn.\,(\ref{eq:token-level-prob}), it is easy to see that $\pi_k \leq \pi_{k+1}$. Therefore, our DiffRate with DDP aligns with the fact that less important tokens should have a larger compression probability. 
To make the training and inference consistent, 
%
we convert ${\pi}_k$ into a $0$-$1$  mask as given by,
\begin{equation}\label{eq:token_level_mask}
{m}_k = \left\{
\begin{aligned}
& 0, \pi_k \geq \alpha, \\
& 1, \pi_k < \alpha,
\end{aligned}
\right.
\end{equation}
where ${m}_k=1$ indicates that the $k$-th token is preserved, and vice versa. 
%

In each vision transformer block, we instantiate two independent re-parameterization modules to learn both pruning and merging compression rates. Thus,  it generates two token-level masks, namely the pruning mask and the merging mask, denoted $m^{p}_k$ and $m^{m}_k$ for each token, respectively. Note that the token removed in last block must also be compressed in this block. Hence, the final mask is defined as 
\begin{equation}\label{eq:final_mask}
    {m}_k = {m}_k \cdot {m}^{p}_{k} \cdot {m}^{m}_{k},
\end{equation}
where $m_k$ on the right-hand side is the mask for the $k$-th token in the last block.
%
%

\textbf{Attention Masking.} 
To preserve the gradient back-propagation chain, we convert token dropping into attention masking with mask $m_k$ in Eqn.\,(\ref{eq:final_mask}) following DynamicViT \cite{rao2021dynamicvit}. 
To achieve this, we construct an attention mask $\mathbf{M}$ with the same dimensions as the attention map for each self-operation operation:
\begin{equation}
{M}_{i,j} = \left\{
\begin{aligned}
& 1, i=j, \\
& {m}_j, i \neq j.
\end{aligned}.
\right. \label{eq:mask_for_attention}
\end{equation}
The attention mask prevents the interaction between all compressed tokens and the other tokens, except itself. We then use this mask to modified the $\mathrm{Softmax}$ operation in the next self-attention module: 
\begin{gather}
\mathbf{S} = \frac{\mathbf{Q} \mathbf{K}^T}{\sqrt{D}}, 
\hat{S}_{i,j} = \frac{\mathrm{exp}({S}_{i,j}){M}_{i,j}}{\sum_{k=1}^N \mathrm{exp}(\mathrm{{S}_{i,k}}){M}_{i,k}},\label{eq:attention_masking}
\end{gather}
where $\mathbf{Q}\in\mathbb{R}^{N\times D}$ is the query matrix, $\mathbf{S}\in\mathbb{R}^{N\times N}$ is the original attention map before $\mathrm{SoftMax}$, and $\hat{S}_{i,j}$ is actually used to update tokens. Eqn.\,(\ref{eq:mask_for_attention}-\ref{eq:attention_masking}) enables propagation the loss function's gradient onto the mask $m$.

%
\section{Training Objective}
We solve the optimization problem in Eqn.\,(\ref{eq:loss}-\ref{eq:forward}) by minimizing the total loss as given by
\begin{equation}\label{eq:objective}
    \mathcal{L} = \mathcal{L}_{cls} + \lambda_{f} \mathcal{L}_{f}(\bm{\alpha}_p, \bm{\alpha}_m),
\end{equation}
where $\mathcal{L}_{f} = (\mathcal{F}(\bm{\alpha}_p, \bm{\alpha}_m)-T)^2$ is the loss to constraint the FLOPs. The hyper-parameter $\lambda_{f}$ balances the two loss terms, and we set it to $5$ by default in our experiments.

During network back-propagation, we utilize the straight-through-estimator (STE)~\cite{hinton2012neural} to calculate the gradient of Eqn.\,(\ref{eq:mask_for_attention}). Hence, we can calculate the gradient of $\mathcal{L}$ with respect to ${\rho}_k$ using the chain rule:
\begin{equation}\label{eq:gradient}
  \frac{\partial \mathcal{L}}{\partial {\rho}_k} =
   \sum_{j=1}^{N}  \frac{\partial \mathcal{L}}{\partial {m_j}} 
  \frac{\partial {m}_j}{\partial {\pi}_j}
  \frac{\partial {\pi}_j}{\partial {\rho}_k} \approx
  \sum_{j=1}^{N} \frac{\partial \mathcal{L}}{\partial {m}_j} 
  \frac{\partial {\pi}_j}{\partial {\rho}_k}.
\end{equation}
Since $\rho_k$ is differentiable through  Eqn.\,(\ref{eq:gradient}), the compression rate $\alpha$ can be optimized with gradient back-propagation by Eqn.\,(\ref{eq:expected_keeping_ratio}). 

\begin{algorithm}[!t]
\caption{Differentiable Compression Rate.}\label{alg:DiffRate}
\textbf{Input}: training dataset $(\mathbf{X},\mathbf{Y})$, pretrained model weight $\mathbf{W}^{*}$, target FLOPs $T$,  DDP with discrete compression rate \scalebox{0.8}{$\{C_k\}_{k=1}^{N}$} and learnable probabilities \scalebox{0.8}{$\{\rho_k\}_{k=1}^{N}$}. \\
\textbf{Output}: block-wise pruning compression rate \scalebox{0.8}{$\bm{\alpha}_{p}=\{\alpha_{p}^{l}\}_{l=1}^L$} and merging compression rate \scalebox{0.8}{$\bm{\alpha}_{m}=\{\alpha_{m}^{l}\}_{l=1}^L$}.
\begin{algorithmic}[1]
\For{$(\mathbf{x},\mathbf{y})$ in $(\mathbf{X},\mathbf{Y})$}
\State calculate \scalebox{0.8}{$\bm{\alpha}_p$} and \scalebox{0.8}{$\bm{\alpha}_m$} with \scalebox{0.8}{$\{\rho_k\}_{k=1}^{N}$} by Eqn.\,(\ref{eq:expected_keeping_ratio}).
\State calculate pruning mask \scalebox{0.8}{$\{m^{p}_{k}\}_{k=1}^N$} by Eqn.\,(\ref{eq:token_level_mask}).
\State calculate merging mask \scalebox{0.8}{$\{m^{m}_{k}\}_{k=1}^N$} by Eqn.\,(\ref{eq:token_level_mask}).
\State sorting $\to$ pruning $\to$ merging in Sec.\,\ref{sec:token-sort}
\State attention masking with Eqn.\,(\ref{eq:mask_for_attention}-\ref{eq:attention_masking})
\State calculate classification loss: \scalebox{0.9}{$\mathcal{L}_{cls}(\mathbf{W}^*(\mathbf{x}),\mathbf{y})$}
\State calculate FLOPs loss: \scalebox{0.9}{${L}_{f} = (\mathcal{F}(\bm{\alpha}_p, \bm{\alpha}_m)-T)^2$} 
\State calculate optimization objective: \scalebox{0.8}{$\mathcal{L} = \mathcal{L}_{cls} + \lambda_{f} \mathcal{L}_{f}$} 
\State backward $\mathcal{L}$ to $\rho_k$ with Eqn.\,(\ref{eq:gradient})
\State update $\rho_k$ by gradient
\EndFor
\State \textbf{return} $\bm{\alpha}_p$ and $\bm{\alpha}_m$
\end{algorithmic}
\end{algorithm}

\textbf{Overall Algorithm.}
The overall training algorithm of DiffRate is illustrated in Algorithm \ref{alg:DiffRate}. It consists of three steps: forward model with $\rho_k$ (Lines 2-6), calculating optimization objective (Lines 7-9), backward propagation and $\rho_k$ update in DDP  (Lines 10-11).
The DiffRate algorithm finds the optimal compression rate by updating $\rho_k$ in a differentiable form, and the resulting compression rate can be directly applied to off-the-shelf models.




\textbf{Extension to Other Complexity Metrics.}
Our DiffRate model offers flexible differentiable compression rates that can be supervised using various computational complexity metrics, such as FLOPs and latency. To investigate its potential, we employed Gemmini~\cite{gemmini}, a framework for generating deep learning accelerators that can produce a diverse set of realistic accelerators based on a flexible architecture template. Using Gemmini, we conducted a co-search of the design space for compression ratios $\bm{\alpha}_p$ and $\bm{\alpha}_m$, as well as accelerator parameters $\bm{\beta}$ simultaneously. The loss function in Eqn.\,(\ref{eq:objective}) becomes
\begin{equation}\label{eq:many-objective}
\begin{split}
    \mathcal{L} &= \mathcal{L}_{cls} + \lambda_{f} \mathcal{L}_{f}(\bm{\alpha}_p, \bm{\alpha}_m)\\
    &+ \lambda_{la} \mathcal{L}_{la}(\bm{\alpha}_p, \bm{\alpha}_m,\bm{\beta}) +\lambda_{pw} \mathcal{L}_{pw}(\bm{\alpha}_p, \bm{\alpha}_m,\bm{\beta}),
\end{split}
\end{equation}
where $\mathcal{L}_{la}$ and $\mathcal{L}_{pw}$ are loss functions to constraint latency and power consumption in Gemmini accelerator, respectively. $\lambda_{la}$ and $\lambda_{pw}$ are their strengths. By minimizing $\mathcal{L}$, DiffRate can further learn optimal $\bm{\alpha}_p^*, \bm{\alpha}_m^*$ satisfying various resource constraints, and the optimum $\bm{\beta}^*$ can be implemented in FPGA board. The details are in Appendix\,\ref{sec:computation_constraint}.

\begin{figure*}[!t]
  \centering
    \vspace{-1em}
  \begin{subfigure}{0.3\linewidth}
    \includegraphics[width=\linewidth]{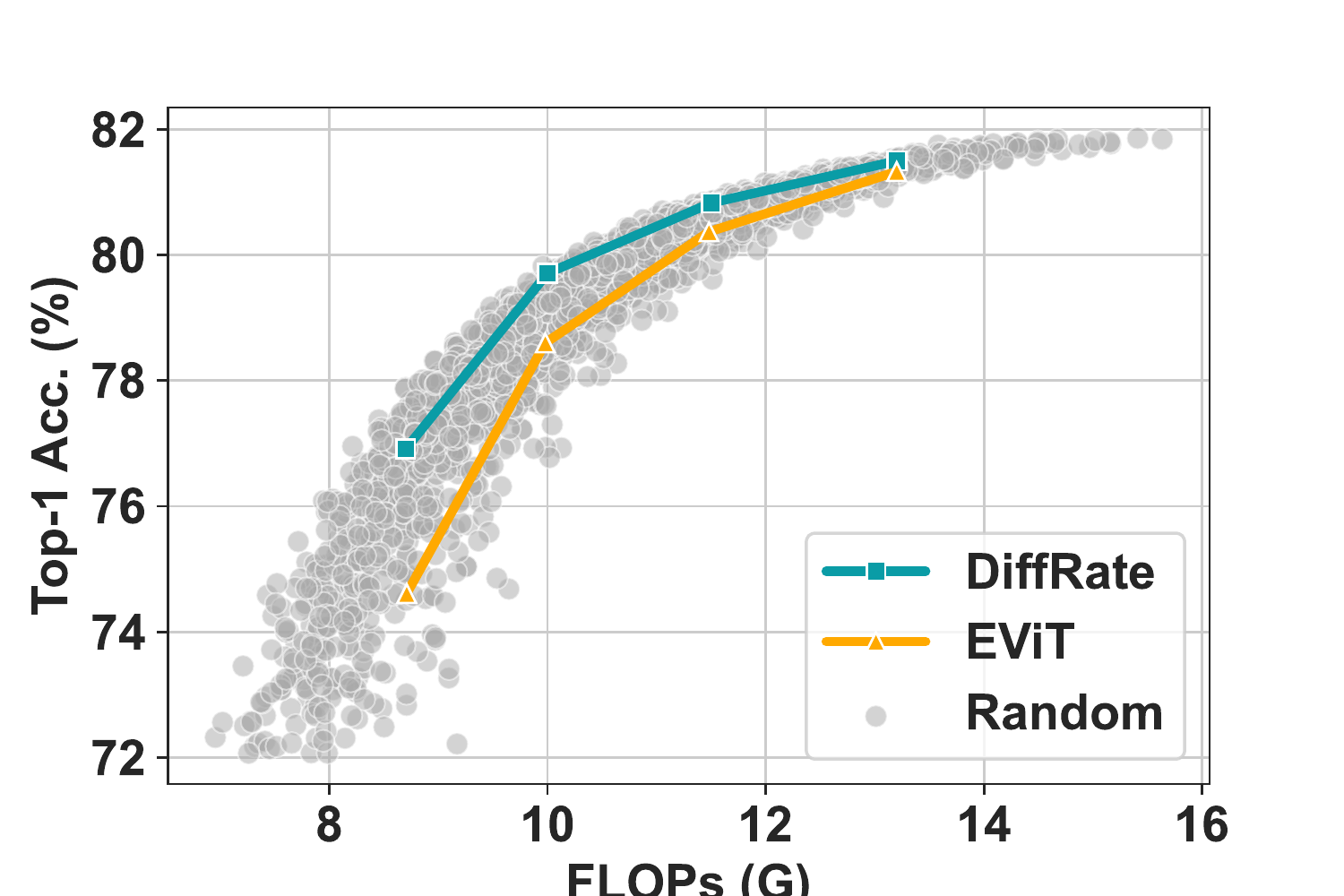}
    \caption{Pruning}
    \label{fig:top12}
  \end{subfigure}
  \hspace{0.13em}
  \begin{subfigure}{0.3\linewidth}
    \includegraphics[width=\linewidth]{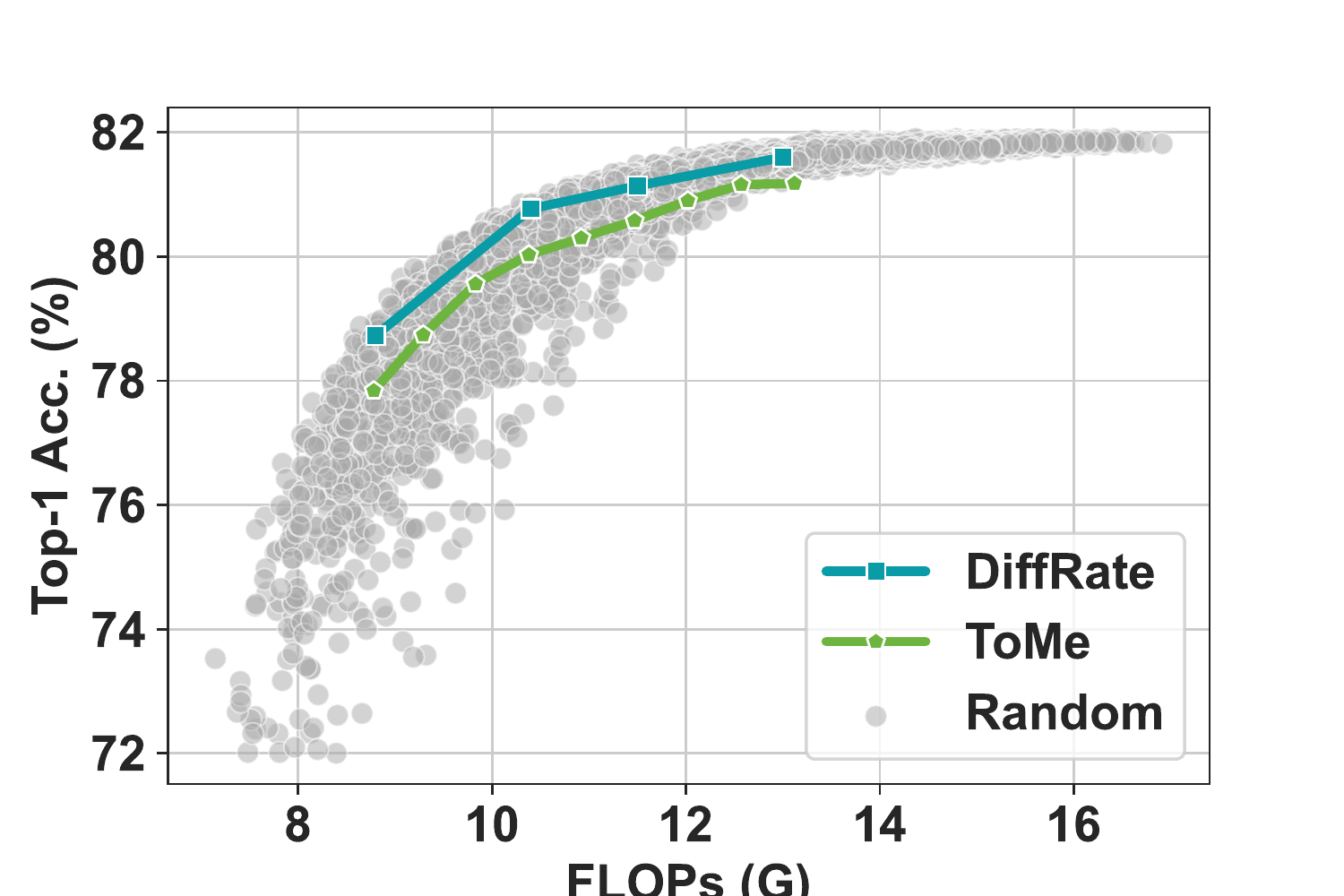}
    \caption{Merging}
    \label{fig:visual_origin}
  \end{subfigure}
  \begin{subfigure}{0.3\linewidth}
    \includegraphics[width=\linewidth]{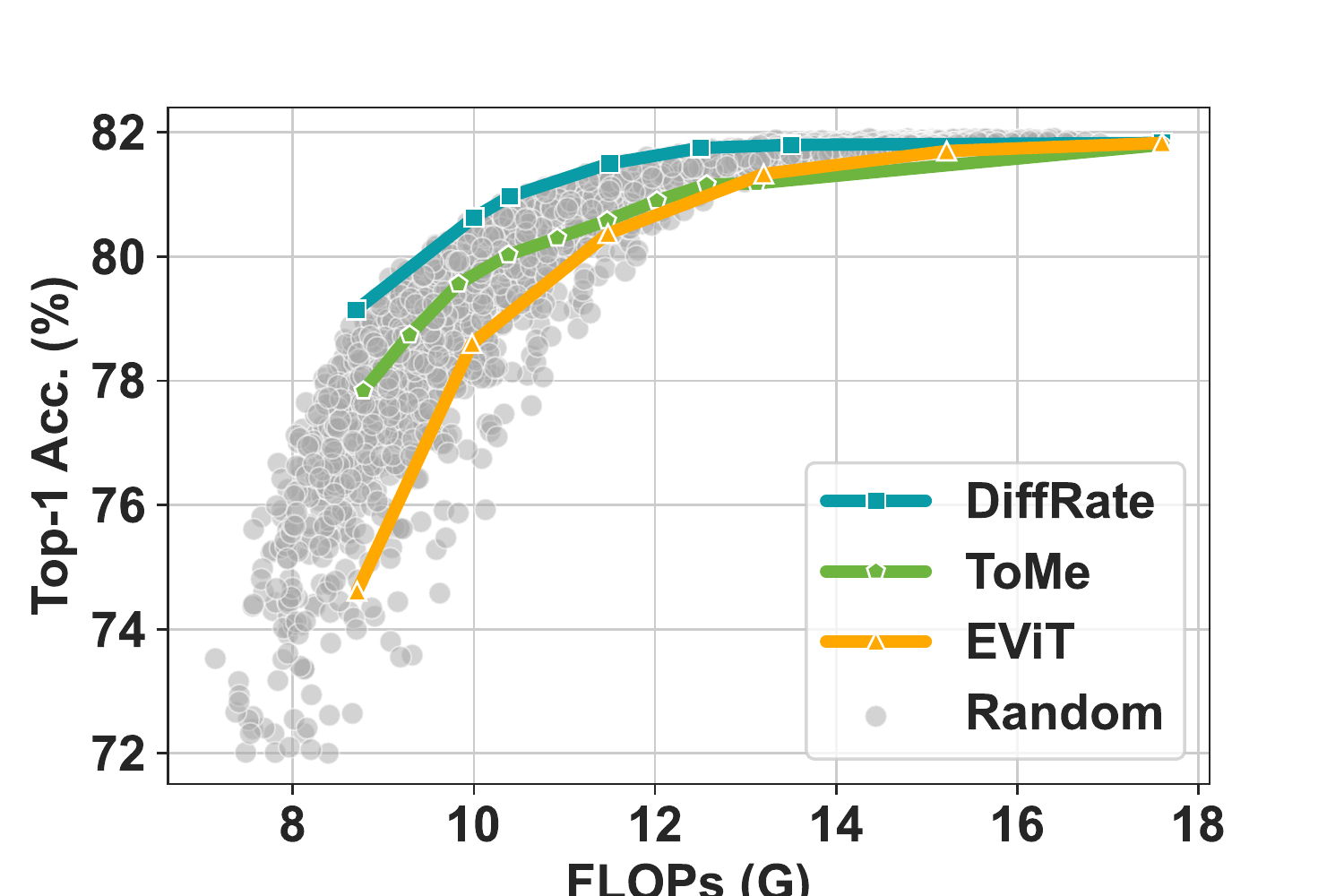}
    \caption{Pruning and Merging}
    \label{fig:visual_mixed}
  \end{subfigure}
  \vspace{-1em}
  \caption{\textbf{Token Compression Schedule Comparison.} We test the proposed DiffRate on ViT-B (DeiT) with three token compression options: (a) Only token pruning like EViT~\cite{zeng2022not}, (b) Only token merging like ToMe~\cite{bolya2022tome},(c) Pruning and Merging as depicted in Sec.\,\ref{sec:token-sort}. The token compression schedules searched through DiffRate outperform the constant schedules used in ToMe~\cite{bolya2022tome} and EViT~\cite{liang2022not}. Moreover, the performance of our method is close to optimal when compared to $10,000$ randomly sampled schedules.}
  \label{fig:schedule_comparison}
  \vspace{-1em}
\end{figure*}

\section{Experiments}
This section presents extensive experiments to verify our proposed DiffRate. Sec. \ref{sec:exp-detail} provides training details. Sec. \ref{sec:exp-performance} compares DiffRate with previous compression techniques. The ablation study and visualization are presented in Sec. \ref{sec:exp-ablation} and Sec. \ref{sec:exp-vis}, repsectively.

\subsection{Implementation Details}\label{sec:exp-detail}
In this section, we conduct a series of experiments on ImageNet-1k~\cite{deng2009imagenet} using  DeiT~\cite{touvron2021training}, MAE~\cite{he2022masked}, and LV-ViT~\cite{jiang2021all}.
We initialize the backbone models with pre-trained models and fix them to train the learnable probabilities in DDP with the objective function in Eqn.\,(\ref{eq:objective}). The DDP is trained for $3$ epochs using a learning rate of $1e^{-2}$. Once the compression rate of DDP is determined, it can be directly applied to off-the-shelf models. We also fine-tune the model optionally for 30 epochs using a learning rate of $1e^{-5}$ after determining the compression rate, denoted by the superscript ``$\dag$'' in the table for differentiation.
All throughput measurements are taken during inference on an A100 GPU with a batch size of $1024$ and fp16. The other experimental setups follow most of the training techniques used in DeiT~\cite{touvron2021training}, and additional details can be found in Appendix\,\ref{sec:training_details}.
Note that our DiffRate is highlighted in the tables in \hl{gray}, and \textbf{bold} denotes the best results.

\begin{table}[!ht]
    \centering
      \caption{\textbf{Token compression on the off-the-shelf models.} We directly apply EViT~\cite{liang2022not}, ToMe~\cite{bolya2022tome}, and the proposed DiffRate to off-the-shelf models, \ie without updating the network parameters of pre-trained models. }
    \vspace{-1em}
    \label{tab:off_the_shelf}
    \setlength\tabcolsep{6pt}
    \small
    \begin{tabular}{ccccc}
    \hline
    Model & Method & FLOPs & imgs/s & Acc. \\
    \hline
     \multirow{7}{*}{\shortstack{ViT-S (DeiT)}}  
              &  Baseline~\cite{touvron2021training}  & 4.6  & 5039  & 79.82   \\
              &  EViT~\cite{liang2022not} & 2.3  & 8950  &  73.83 \\
              &  ToMe~\cite{bolya2022tome} & 2.3  & 8874  &  77.99 \\
    & \cellcolor{Gray}\textbf{DiffRate} &\cellcolor{Gray}  \textbf{2.3} &\cellcolor{Gray}  8901  &\cellcolor{Gray}  \textbf{78.76} \\
              &  EViT~\cite{liang2022not} & 3.0  & 6807  &  78.50 \\
              &  ToMe~\cite{bolya2022tome} & 2.9  & 6712  &  78.89 \\
   & \cellcolor{Gray}  \textbf{DiffRate} & \cellcolor{Gray} \textbf{2.9} &\cellcolor{Gray}  6744  &\cellcolor{Gray}  \textbf{79.58} \\
    \hline
    \multirow{7}{*}{ViT-B (DeiT)}  
              &  Baseline~\cite{touvron2021training}  & 17.6  & 2130  & 81.83   \\
              &  EViT~\cite{liang2022not} & 8.7  & 4230  & 74.61    \\
              &  ToMe~\cite{bolya2022tome} &  8.8 & 4023  & 77.84  \\
       & \cellcolor{Gray}  \textbf{DiffRate}  & \cellcolor{Gray}\textbf{8.7}  &\cellcolor{Gray} 4124  & \cellcolor{Gray}\textbf{78.98}  \\
              &  EViT~\cite{liang2022not} & 11.5  & 2886  & 80.37    \\
              &  ToMe~\cite{bolya2022tome} &  11.5 & 2834  & 80.58  \\
       &\cellcolor{Gray}   \textbf{DiffRate}  &\cellcolor{Gray} \textbf{11.5}  &\cellcolor{Gray} 2865  &\cellcolor{Gray} \textbf{81.50}  \\
    \hline
    \multirow{7}{*}{ViT-B (MAE)} 
              &  Baseline~\cite{he2022masked}  & 17.6  & 2130  & 83.72   \\
              &  EViT~\cite{liang2022not} & 8.7  & 4230  & 75.15    \\
              &  ToMe~\cite{bolya2022tome}  & 8.8  & 4023 & 78.86    \\ 
    &\cellcolor{Gray}   \textbf{DiffRate}  &\cellcolor{Gray} \textbf{8.7}  &\cellcolor{Gray} 4150  &\cellcolor{Gray}  \textbf{79.96} \\
              &  EViT~\cite{liang2022not} & 11.5  & 2886  & 82.01    \\
              &  ToMe~\cite{bolya2022tome}  & 11.5  & 2834 & 82.32    \\ 
    &\cellcolor{Gray}   \textbf{DiffRate}  &\cellcolor{Gray} \textbf{11.5}  &\cellcolor{Gray} 2865  &\cellcolor{Gray}  \textbf{82.91} \\
    \hline
    \multirow{7}{*}{ViT-L (MAE)} 
              &  Baseline~\cite{he2022masked}  & 61.6  & 758  & 85.95   \\
              &  EViT~\cite{liang2022not} & 29.7  & 1672  & 81.52    \\
              &  ToMe~\cite{bolya2022tome}  & 31.0  & 1550 & 84.24    \\ 
    &\cellcolor{Gray}   \textbf{DiffRate}  &\cellcolor{Gray} \textbf{31.0}  &\cellcolor{Gray} 1580  &\cellcolor{Gray} \textbf{84.65}  \\
              &  EViT~\cite{liang2022not} & 39.6  & 1089  & 85.06    \\
              &  ToMe~\cite{bolya2022tome}  & 42.3  & 1033 & 85.41    \\ 
    &\cellcolor{Gray}   \textbf{DiffRate}  &\cellcolor{Gray} \textbf{42.3}  &\cellcolor{Gray} 1045  &\cellcolor{Gray} \textbf{85.56}  \\
    \hline
    \multirow{7}{*}{ViT-H (MAE)} 
              &  Baseline~\cite{he2022masked}  & 167.4  & 299  & 86.88   \\
              &  ToMe~\cite{bolya2022tome}  &  92.9 & 500 &  86.01   \\ 
              &  EViT~\cite{liang2022not} & 99.1  & 512  & 85.54    \\
     &\cellcolor{Gray}   \textbf{DiffRate} &\cellcolor{Gray} \textbf{93.2}  &\cellcolor{Gray}  504 &\cellcolor{Gray}  \textbf{86.40}  \\
              &  ToMe~\cite{bolya2022tome}  &  103.4 & 442 &  86.29   \\ 
              &  EViT~\cite{liang2022not} & 112.9  & 432  & 86.32    \\
    &\cellcolor{Gray}   \textbf{DiffRate} &\cellcolor{Gray} \textbf{103.4}  &\cellcolor{Gray}  450 &\cellcolor{Gray}  \textbf{86.72}  \\   
    \hline
    \multirow{3}{*}{LV-ViT-S} 
              &  Baseline~\cite{jiang2021all}  & 6.6  & 3630  & 83.30   \\
              &  EViT~\cite{liang2022not}  &  3.9 & 5077 &  79.77   \\ 
&\cellcolor{Gray}   \textbf{DiffRate} &\cellcolor{Gray} \textbf{3.9}  &\cellcolor{Gray} 5021 &\cellcolor{Gray} \textbf{82.56}  \\
    \hline
    \end{tabular}
    \vspace{-1em}
\end{table}

\subsection{Comparison with state-of-the-art} \label{sec:exp-performance}

\textbf{Compression Schedule.} 
In Fig.\,\ref{fig:schedule_comparison}, we compare the compression rate obtained by DiffRate with three other token pruning schedules, including EViT~\cite{zeng2022not}, ToMe~\cite{bolya2022tome}, and randomly sampled schedules. We evaluate the FLOPs and accuracy on the ImageNet-1k validation dataset using an off-the-shelf ViT-B (DeiT) and investigate three token compression options: only pruning, only merging, and a combination of pruning and merging, as depicted in Sec.\,\ref{sec:token-sort}. We can observe that DiffRate performed almost optimally, regardless of the token compression setting and FLOPs constraint. Additionally, DiffRate's advantage was more prominent at lower FLOPs constraints, indicating its ability to provide more appropriate compression rates at larger solution space. DiffRate also benefited from unified token compression, indicating that it can preserve more information by combining the merits of token merging and pruning.
%

\textbf{Off-the-shelf.} 
We compare the proposed DiffRate with EViT~\cite{zeng2022not} and ToMe~\cite{bolya2022tome} in the ``off-the-shelf'' setting, indicating direct use of the pre-trained model without updating the network parameters. 
Notably, the compression rate achieved by DiffRate can be applied to pre-trained models without any additional computation. 
As shown in Table\,\ref{tab:off_the_shelf}, DiffRate consistently outperforms both EViT~\cite{liang2022not} and ToMe~\cite{bolya2022tome} across various models. Specifically, DiffRate outperforms ToMe~\cite{bolya2022tome} by 0.14\% to 1.14\%, and outperforms EViT~\cite{liang2022not} by 0.39\% to 4.93\%. 
These results demonstrate the strong potential of DiffRate as an effective post-training token compression method.
More results at other FLOPs constraints can be found in Appendix\,\ref{sec:more_results}.
%

\begin{table}[!t]
    \centering
        \caption{\textbf{Token compression with training.} $^{\dag}$ indicates fine-tuning the model with searched compression rate for 30 epochs. }
    \vspace{-1em}
    \label{tab:with_train}
    \setlength\tabcolsep{5pt}
    \small
    \begin{tabular}{ccccc}
    \hline
    Model & Method & FLOPs & imgs/s & Acc. \\
    \hline
    \multirow{7}{*}{ViT-S (DeiT)}  
              &  Baseline~\cite{touvron2021training}  & 4.6  & 5039  & 79.82   \\
              &  DynamicViT~\cite{rao2021dynamicvit}  & 2.9  & 6527  & 79.30   \\
              &  Evo-ViT~\cite{xu2022evo} & 3.0  & 6679 & 79.40  \\
              &  EViT~\cite{liang2022not} & 3.0  & 6807  &  79.50 \\
              &  ToMe~\cite{bolya2022tome}  & 2.9 & 6712 &  79.49   \\ 
              &  ATS~\cite{fayyaz2022ats} & 2.9  & - & 79.70  \\
              &  SPViT~\cite{kong2022spvit} &  2.6 & -  & 79.34  \\
             &\cellcolor{Gray}   \textbf{DiffRate} &\cellcolor{Gray} \textbf{2.9}  &\cellcolor{Gray}  6744 &\cellcolor{Gray} \textbf{79.58}  \\
             &\cellcolor{Gray}   \textbf{DiffRate}$^\dag$ &\cellcolor{Gray} \textbf{2.9}  &\cellcolor{Gray}  6744 &\cellcolor{Gray} \textbf{79.83}  \\
    \hline
    \multirow{4}{*}{ViT-B (DeiT)}  
              &  Baseline~\cite{touvron2021training}  & 17.6  & 2130  & 81.83   \\
              &  EViT~\cite{liang2022not} & 11.5  & 2886  & 81.30    \\
              &  ToMe~\cite{bolya2022tome} & 11.5  & 2834 & 81.41  \\
              &\cellcolor{Gray}   \textbf{DiffRate} &\cellcolor{Gray} \textbf{11.5}  &\cellcolor{Gray} 2865  &\cellcolor{Gray} \textbf{81.50}  \\
              &\cellcolor{Gray}   \textbf{DiffRate}$^\dag$ &\cellcolor{Gray} \textbf{11.5}  &\cellcolor{Gray} 2865  &\cellcolor{Gray} \textbf{81.71}  \\
    \hline
    \multirow{4}{*}{ViT-B (MAE)} 
              &  Baseline~\cite{he2022masked}  & 17.6  & 2130  & 83.72   \\
              &  ToMe~\cite{bolya2022tome}  & 11.5  & 2834 &  82.94   \\ 
              &\cellcolor{Gray}   \textbf{DiffRate} &\cellcolor{Gray} \textbf{11.5}  &\cellcolor{Gray} 2865  &\cellcolor{Gray} \textbf{82.91}  \\
              &\cellcolor{Gray}   \textbf{DiffRate}$^\dag$ &\cellcolor{Gray} \textbf{11.5}  &\cellcolor{Gray} 2865  &\cellcolor{Gray} \textbf{83.25}  \\
    \hline
    \multirow{4}{*}{ViT-L (MAE)} 
              &  Baseline~\cite{he2022masked}  & 61.6  & 758  & 85.95   \\
              &  ToMe~\cite{bolya2022tome}  & 42.3  & 1033 & 85.59    \\ 
              &\cellcolor{Gray}   \textbf{DiffRate} &\cellcolor{Gray} \textbf{42.3}  &\cellcolor{Gray} 1045  &\cellcolor{Gray} \textbf{85.56}  \\
              &\cellcolor{Gray}   \textbf{DiffRate}$^\dag$ &\cellcolor{Gray} \textbf{42.3}  &\cellcolor{Gray} 1045  &\cellcolor{Gray} \textbf{85.71}  \\
    \hline
    \multirow{3}{*}{ViT-H (MAE)} 
              &  Baseline~\cite{he2022masked}  & 167.4  & 299  & 86.88   \\
              &  ToMe~\cite{bolya2022tome}  & 103.4  & 442 & 86.51    \\ 
              &\cellcolor{Gray}   \textbf{DiffRate} & \cellcolor{Gray}\textbf{103.4}  &\cellcolor{Gray} 450  &\cellcolor{Gray} \textbf{86.72}  \\
    
    \hline
    \end{tabular}
    \vspace{-1em}
\end{table}

\textbf{With Training.} 
In Table\,\ref{tab:with_train}, we compare the proposed DiffRate with several methods that require training the model, indicating fine-tuning with pre-trained models or training from scratch. The evaluated methods include EViT~\cite{bolya2022tome} and ToMe~\cite{bolya2022tome}, which train models from scratch, and ATS~\cite{fayyaz2022ats}, DynamicViT~\cite{rao2021dynamicvit}, SP-ViT~\cite{kong2022spvit}, which fine-tune pre-trained models for 30 epochs. 
To ensure a fair comparison, for DiffRate, we also fine-tune the pre-trained models for 30 epochs with the searched compression rate. 
We can observe that DiffRate still maintains a performance advantage compared to methods that require training. Specifically, ATS achieves a top-1 accuracy of 79.70\% in DeiT-S, close to our 79.83\%. However, it is important to note that ATS is an input-adaptive token pruning method that cannot be applied to batch inference, while DiffRate does not suffer this problem. 
Moreover, we find that DiffRate utilized with an off-the-shelf model achieves comparable or superior performance to methods that require training. For instance, DiffRate attains 79.58\% on the off-the-shelf DeiT-S~\cite{touvron2021training}, while EViT and ToMe, after training, only achieve 79.50\% and 79.49\%, respectively. Similar results are also observed in ViT-B (DeiT) and ViT-B/L/H (MAE).

\begin{table}[!t]
    \centering
    \setlength\tabcolsep{2.6pt}
    \caption{\textbf{Results under multiple complexity constraints.} on ViT-S (DeiT). DiffRate indicates single FLOPs constraint as Eqn.\,(\ref{eq:objective}), DiffRate-M indicates multiple complexity constraint as Eqn.\,(\ref{eq:many-objective}).}
    \vspace{-1em}
    \begin{tabular}{ccccc}
    \hline
    Method & FLOPs(G) & Latency(ms) & Power(mW) & Acc.  \\
    \hline
    Baseline & 4.6 & 68.1 & 156 & 79.82 \\
    EViT & 3.0 & 40.4 & 99 & 79.50 \\
    DiffRate & 2.9 & 40.1 & 98 & \textbf{79.83} \\
    \rowcolor[gray]{0.9} \textbf{DiffRate-M} &  \textbf{2.9} &  \textbf{37.6} &  \textbf{90} &  79.80 \\
    \hline
    \end{tabular}
    \label{tab:multl-objective}
    \vspace{-0.5em}
\end{table}

\begin{table*}[!ht]
  \caption{\textbf{Ablation experiments} using ViT-B (DeiT)~\cite{touvron2021training}. Our default settings are marked in gray.}
    \vspace{-1em}
  \centering
  \captionsetup[subtable]{position=bottom}
  \subcaptionbox{\textbf{Sorting Metric.} Simply class attention can measure the importance of tokens.\label{tab:sorting_metric}}{
    \setlength\tabcolsep{3pt}
    \begin{tabular}{cc}
    \hline
    Metric  & Acc.(\%) \\
    \hline
    Random & 79.72 \\
    $\mathbf{A_{i}}$ & 81.38 \\
    $\mathbf{A_{c} \cdot |\mathbf{V}|}$ & \textbf{81.53} \\
    \rowcolor[gray]{0.9}  $\mathbf{A_{c}}$ & 81.50 \\
    \hline
    \end{tabular}
  }
  \quad
  \subcaptionbox{\textbf{Token Compression Module Option.} A merging and pruning pipeline is the best choice.\label{tab:token_compression_option}}{
  \setlength\tabcolsep{3pt}
    \begin{tabular}{cc}
    \hline
    Option &  Acc.(\%) \\
    \hline
    Pruning  &  80.83 \\
    Merging  & 81.14 \\
    Merging-Pruning & 81.18 \\
    \rowcolor[gray]{0.9}   Pruning-Merging & \textbf{81.50} \\
    \hline
    \end{tabular}
  }
  \quad
  \subcaptionbox{\textbf{Training Data.}\label{tab:training_data} 1,000 images is enough to optimize compression rate.}{
  \setlength\tabcolsep{3pt}
    \begin{tabular}{cc}
    \hline
    Number & Acc.(\%) \\
    \hline
    1,000  & 81.40 \\
    4,000  &  81.46 \\
    16,000 & \textbf{81.50} \\
    \rowcolor[gray]{0.9}    All & \textbf{81.50} \\
    \hline
    \end{tabular}
  }
  \quad
  \subcaptionbox{\textbf{Optimization Time.}\label{tab:optimazation_time} Token compression rate can converge within 2.7 gpu houres.}{
  \setlength\tabcolsep{3pt}
    \begin{tabular}{cc}
    \hline
    Time (g-hrs) &  Acc.(\%) \\
    \hline
    0.9 (1-ep)  &  81.32 \\
   \rowcolor[gray]{0.9} 2.7 (3-ep)  &  81.50 \\
    9 (10-ep)  &  81.50 \\
    27  (30-ep)  &  \textbf{81.52} \\
    \hline
    \end{tabular}
  }
  \label{fig:experimental_results}
  \vspace{-0.5em}
\end{table*}

\begin{figure*}[!ht]
    \centering
    \includegraphics[width=0.9\linewidth]{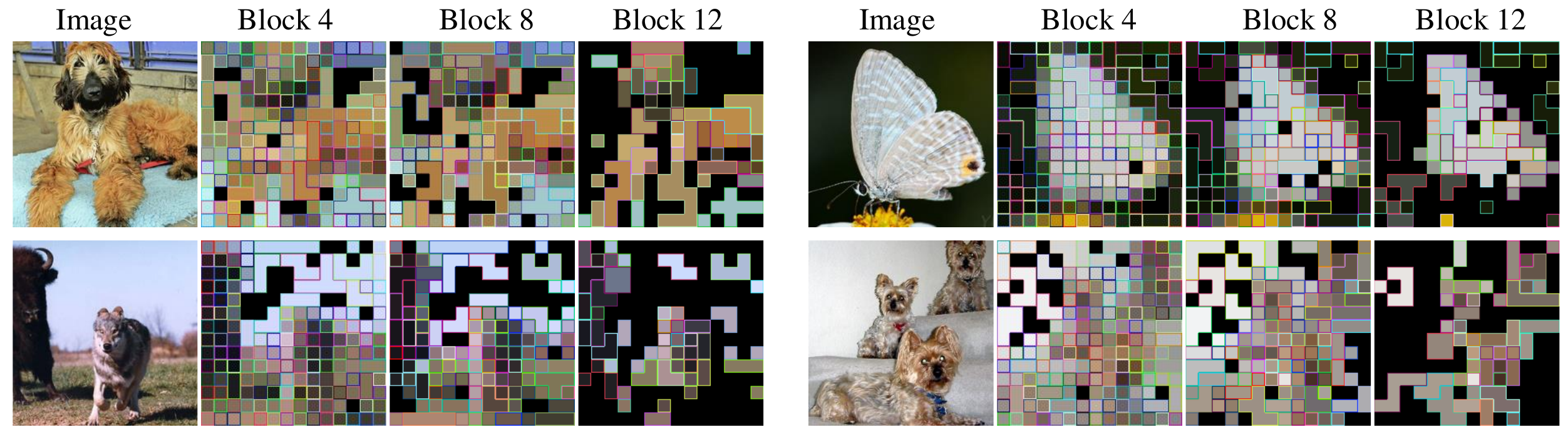}
    \vspace{-1em}
    \caption{\textbf{Image Visualizations.} Results of proposed DiffRate on ImageNet-1k validation using a pre-trained DeiT-B model, only 34 tokens left in block 12. Merged tokens are represented by patches with the same inner and border color, while pruned tokens are represented by black. The visualizations show that DiffRate can gradually prune the redundant token in the background and merge less-discriminative tokens in the foreground.}
    \label{fig:visualization}
    \vspace{-1em}
\end{figure*}

\textbf{Multiple Complexity Constraints.} 
In addition, we supervise DiffRate with three computational complexity constraints: FLOPs, latency, and power, as detailed in Eqn.\,(\ref{eq:many-objective}). As shown in Table\,\ref{tab:multl-objective}, integrating multiple complexity constraints enhances the trade-off between performance and complexity. Specifically, leveraging multiple complexity constraints leads to a remarkable reduction of 2.5ms in latency and 8mW in power consumption compared to a single FLOPs constraint, with only a negligible decline in accuracy of a mere 0.03\%.

\subsection{Ablation Study}\label{sec:exp-ablation}
In this section, we perform DiffRate with variants to investigate the effectiveness of our proposed method.
As the experiment shown in Table\,\ref{tab:sorting_metric}, we compare the influence of several token sorting metrics, including randomly generated rank, class attention ($A_{c}$), image attention ($A_{i}$)~\cite{liang2022not}, and  the product of class attention and the value matrix's norm ($A_{c}\cdot|V|$)~\cite{fayyaz2022ats}. It can be observed that $A_{c}$ and $A_{c}\cdot|V|$ exhibit similar performance, and we choose $A_{c}$ as our default setting since it does not require any additional computation.
Then Table\,\ref{tab:token_compression_option} compares several token compression options, including only pruning, only merging, pruning then merging, and merging then pruning. The results show that pruning then merging performs the best since it successfully combines the advantages of both pruning and merging. 
What's more, in Table\,\ref{tab:training_data}, we investigate the amount of training data required by DiffRate to find the optimal compression rate. Surprisingly, we find that only $1,000$ images are sufficient to obtain an appropriate compression rate. Although we also optimize the token compression rate using the entire training dataset, our results demonstrate the potential for DiffRate to work well even with minimal data.
Lastly, Table\,\ref{tab:optimazation_time} investigates the convergence time required by DiffRate. We can find that only three epochs are required, demonstrating the efficiency of DiffRate as a token compression approach.


\subsection{Visualization}\label{sec:exp-vis}

%
The visualizations results in Fig.\,\ref{fig:visualization} demonstrate that DiffRate  effectively removes semantically irrelevant background information. 
Furthermore, DiffRate can reduce the number of tokens by merging less-discriminative tokens in the foreground. For example, in the first row, DiffRate successfully removes most of the background and merges the dog hair and butterfly wings tokens into fewer tokens. In the second row, DiffRate preserves salient information tokens in different image regions, even when multiple instances exist.
Overall, the visualization results highlight the effectiveness of our proposed DiffRate method in compressing ViT models without significant information loss.
See more results in Appendix\,\ref{sec:more_visualization}.

\section{Conclusion}
This work presents a new token compression framework, named Differentiable Compression Rate (DiffRate). The proposed approach integrates both token pruning and merging into a unified framework that can optimize the compression rate in a differentiable manner. To achieve this, we introduced a novel Differentiable Discrete Proxy (DDP) module that can effectively determine the optimal compression rate using gradient back-propagation. 
Our experimental results demonstrate that DiffRate can perform comparable or superior to previous state-of-the-art token compression methods, even without fine-tuning the model.
Additionally, DiffRate is highly data-efficient, as it can identify the appropriate compression rate using only $1,000$ images.
Overall, the proposed DiffRate framework offers a new perspective on token compression by revealing the importance of compression rate. We believe that this approach has the potential to pave the way for further advancements in token compression research.


\section*{Acknowledgement}
\begin{sloppypar}
This work is supported by National Key R\&D Program of China
(No.2022ZD0118201) , the National Science Fund for Distinguished Young Scholars (No.62025603), the National Natural Science Foundation of China (No. U21B2037, No. U22B2051, No. 62176222, No. 62176223, No. 62176226, No. 62072386, No. 62072387, No. 62072389, No. 62002305 and No. 62272401), and the Natural Science Foundation of Fujian Province of China (No.2021J01002, No.2022J06001). This work is also partially supported by the National Key R\&D Program of China(NO.2022ZD0160100), and in part by  Shanghai Committee of Science and Technology (Grant No. 21DZ1100100).
\end{sloppypar}

{\small
\bibliographystyle{ieee_fullname}
\bibliography{egbib}
}

\appendix

\newcounter{chapter}
\renewcommand{\thechapter}{\Alph{chapter}}
\renewcommand{\thesection}{\thechapter\arabic{section}} 

\renewcommand{\thesection}{\Alph{section}}
\definecolor{darkgray}{gray}{0.7} 

\section{Training Details}\label{sec:training_details}
Our training receipt follows most of the techniques used in DeiT~\cite{touvron2021training}. In Table\,\ref{tab:training_setting_searching}, we provide the default setting for compression rate searching. In this case, we initialize the backbone with their official pre-trained models, fix the network parameters, and train the proposed Differentiable Discrete Proxy for 3 epochs. In the first epoch, we optionally set $\lambda_f$ in Eq.\,(\textcolor{red}{13}) to 0 for warm-up.
To fine-tune the compressed models, we fix the compression rate and update the network parameters for 30 epochs. 
The default settings for fine-tuning are provided in Table~\ref{tab:training_setting_finetuning}. 
During compression rate searching, we employ attention masking to simulate token dropping, while in the fine-tuning process, we directly drop the redundant tokens. Any differences from the default DeiT recipe are highlighted in \textbf{bold} in the tables.

\begin{table}[!h]
    \centering
    \caption{Compression rate searching training settings.}
    \begin{tabular}{c|c}
    \hline
    cofig & value \\
    \hline
    optimizer & AdamW \\
    \textbf{learning rate} & \textbf{0.01} \\
    \textbf{minimal learning} rate & \textbf{0.001} \\
    learning rate schedule & cosine decay \\
    \textbf{weight decay} & \textbf{0} \\
    batch size & 1024 \\
    \textbf{training epochs} & \textbf{3} \\
    augmentation & RandAug(9, 0.5) \\
    LabelSmoth & 0.1 \\
    DropPath & 0.1 \\
    Mixup & 0.8 \\
    CutMix & 1.0 \\
    \hline
    \end{tabular}
    \label{tab:training_setting_searching}
\end{table}

\begin{table}[!h]
    \centering
    \caption{Fine-tuning training settings.}
    \begin{tabular}{c|c}
    \hline
    cofig & value \\
    \hline
    optimizer & AdamW \\
    \textbf{learning rate} & \textbf{2e-5} \\
    \textbf{minimal learning rate} & \textbf{1e-6} \\
    learning rate schedule & cosine decay \\
    weight decay & 0.05 \\
    batch size & 1024 \\
    \textbf{training epochs} & \textbf{30} \\
    augmentation & RandAug(9, 0.5) \\
    LabelSmoth & 0.1 \\
    DropPath & 0.1 \\
    Mixup & 0.8 \\
    CutMix & 1.0 \\
    \hline
    \end{tabular}
    \label{tab:training_setting_finetuning}
\end{table}

\section{Computational Cost Constraint Details}\label{sec:computation_constraint}
This section provides more details about the computational cost constraint of DiffRate, including traditional FLOPs, and hardware-aware metrics like latency and power consumption.

\subsection{FLOPs Calculation}

\begin{algorithm}[!h]
\caption{FLOPs Calculation.}\label{alg:flop_calculation}
\textbf{Input}:  block-wise pruning compression rate \scalebox{0.9}{$\bm{\alpha}_{p}=\{\alpha_{p}^{l}\}_{l=1}^L$} and merging compression rate \scalebox{0.9}{$\bm{\alpha}_{m}=\{\alpha_{m}^{l}\}_{l=1}^L$}, embedding size $C$. \\
\textbf{Output}: FLOPs $\mathcal{F}(\bm{\alpha}_{p},\bm{\alpha}_{m})$.
\begin{algorithmic}[1]
\State $\alpha^0=0$
\State $\mathcal{F}(\bm{\alpha}_{p},\bm{\alpha}_{m})=0$\Comment{FLOPs}
\For{l=1 to L}
\State $\mathcal{F}(\bm{\alpha}_{p},\bm{\alpha}_{m}) \,\verb|+=|\, 4NC^2 + 2N^2C$\Comment{Attention}
\State $\alpha^l = max(\alpha^{l-1},\alpha_p^l,\alpha_m^l)$
\State $N = N(1-\alpha^l)$
\State $\mathcal{F}(\bm{\alpha}_{p},\bm{\alpha}_{m}) \,\verb|+=|\ 8NC^2$\Comment{MLP}
\EndFor
\State \textbf{return} $\mathcal{F}(\bm{\alpha}_{p},\bm{\alpha}_{m})$
\end{algorithmic}
\end{algorithm}

By utilizing the compression rate obtained from our proposed DiffRate, we can calculate the corresponding FLOPs $\mathcal{F}(\bm{\alpha}_{p},\bm{\alpha}_{m})$ using Algorithm\,\ref{alg:flop_calculation}. The final FLOPs calculation also includes the patch embedding layer and classifier, which are excluded from Algorithm\,\ref{alg:flop_calculation} for clarity. Additionally, we utilize the straight-through estimator (STE) for backpropagation in the $max$ operation of Line 5.

\begin{table}[!th]
    \centering
        \caption{\textbf{Gemmini Search Space.} The Gemmini search space is determined by the number of tiles/meshes in each row and column, which indicates its computational resources, while the bank number/capacity of the scratchpad memory and accumulator determines its memory resources. The buswidth sets an upper limit on the communication speed between the scratchpad memory and computation modules.}
    \vspace{-1em}
    \label{tab:hw}
    \setlength\tabcolsep{2pt}
    \small
    \begin{tabular}{ccc}
    \hline
    parameters & type & search space \\
    \hline
    Tiles in a row & int & 1,2,4,8 \\
    Tiles in a column & int & 1,2,4,8 \\
    Meshes in a row & int & 4,8,16,32 \\
    Meshes in a column & int & 4,8,16,32 \\
    Buswidth (bit) & int & 64,128,256,512 \\
    Bank number of scratchpad memory & int & 1,2,4,8,16 \\
    Capacity of scratchpad memory (MB) & int & 0.25,0.5,1,2,4 \\
    Capacity of accumulator (KB) & int & 64,128,256,512,1024 \\
    \hline
    \end{tabular}
    \vspace{-1em}
\end{table}

\subsection{Latency and Power Constraint}
Apart from considering FLOPs, we can incorporate a differentiable method that accounts for hardware (HW) performance metrics, including latency and power consumption. Both latency and power consumption can be constrained in a similar manner. Therefore, we utilize the notation $\mathcal{L}_{hw}$ to represent both $\mathcal{L}_{la}$ and $\mathcal{L}_{pw}$ in Eq.\,(\textcolor{red}{15}) of the main text. The loss function of a hardware metric is given by:
\begin{gather}\label{eq:loss_hw}
    \mathcal{L}_{hw} = \log(\cosh({E}_{hw}(\bm{\alpha}_{p},\bm{\alpha}_{m},\bm{\beta}) - {T}_{hw}))
\end{gather}
where $\mathcal{L}_{hw}$ represents the corresponding HW performance constraint loss, ${E}_{hw}$ is the HW performance with HW parameters $\bm{\beta}$ and compression rates $\bm{\alpha}_{p},\bm{\alpha}_{m}$, ${T}_{hw}$ is the given target HW performance.


To find the optimal solution that considers both accuracy and HW performance metrics, we can co-explore the design space of compression ratio and HW simultaneously. We use Gemmini~\cite{gemmini}, a deep-learning accelerator generation framework that can produce a wide range of realistic accelerators from a flexible architecture template. The HW parameters $\bm{\beta}$ are provided in Table \ref{tab:hw}. 

We adopt iterative training method to optimize compression rates and HW parameters. To be specific, we first optimize compression rate given HW parameters, and then optimize HW parameters given compression rates. To make the loss of hardware metric differentiable w.r.t. compression rates, we formulate ${E}_{hw}$ as
\begin{equation}\label{eq:loss_alpha}
    E^\alpha_{hw} = \sum_{l=1}^L (\alpha^l + \mathrm{SG}(1-\alpha^l))\mathcal{F}'(\alpha^l,\bm{\beta}^*)
\end{equation}
where $\alpha^l = max(\alpha^{l-1},\alpha_p^l,\alpha_m^l)$, and $SG(\cdot)$ is the operator of stopping gradient. $\mathcal{F}'$ denotes the HW cost model, which can calculate a hardware metric given compression rate and HW parameters. `*' indicates that the HW parameters are fixed.

On the other hand, to make hardware metric differentiable w.r.t. compression rates, we formulate ${E}_{hw}$ as
\begin{equation}\label{eq:loss_beta}
    E^\beta_{hw} = \sum_{h=1}^H\sum_{l=1}^L (\beta_h+ \mathrm{SG}(1-\beta_h))\mathcal{F}'(\alpha^{l*},\bm{\beta})
\end{equation}
where $\beta_h=\mathrm{GS}(\pi_h)$, $\pi_h$ is learnable parameter for $h$-th HW parameter, and GS represents Gumbel-Softmax function.
Here `*' indicates that the compression rates are fixed.

With Eqn.(\ref{eq:loss_alpha}) and Eqn.(\ref{eq:loss_beta}), we can update the compression rate and HW designs in a
differentiable manner to satisfy the target latency and power
consumption.

\section{More Results}\label{sec:more_results}
In this section, we present detailed results from our experiments. Firstly, we present the results obtained across different FLOPs constraints for the DeiT~\cite{touvron2021training} in Sec.~\ref{sec:deit_models} and for the MAE~\cite{he2022masked} in Sec.~\ref{sec:mae_models}. In Sec.~\ref{sec:searched_compression_schedule}, we provide an detailed of the compression schedule that we searched. Additionally, we investigate the transferability of the searched compression schedules across different models in Sec.~\ref{sec:compression_rate_transfer}. Finally, we demonstrate the effectiveness of the compressed models by training them from scratch with faster training speeds in Sec.~\ref{sec:train_from_scratch}.

\subsection{DeiT Models}\label{sec:deit_models}
Table~\ref{tab:full_results_deit} displays the comprehensive results of DeiT~\cite{touvron2021training}, encompassing both off-the-shelf models and fine-tuned models.

\begin{table}[!th]
    \centering
    \caption{\textbf{Full DeiT Results.} FT denotes fine-tuning the compressed model for 30 epochs. \textcolor{darkgray}{Gray} denotes the official un-compressed pre-trained models (Baseline).}
    \begin{tabular}{cccc}
    \hline
    \multirow{2}{*}{Model} & \multirow{2}{*}{FLOPs(G)} & \multicolumn{2}{c}{Acc.(\%)} \\
    & & w/o FT & w/ FT \\
    \hline
    \multirow{6}{*}{\shortstack{ViT-T (DeiT)}}  
    & \textcolor{darkgray}{1.3} & \textcolor{darkgray}{72.13} & \textcolor{darkgray}{-} \\
    & 0.6 & 70.36 & 71.11 \\
    & 0.7 & 71.16 & 71.70 \\
    & 0.8 & 71.74 & 72.18 \\
    & 0.9 & 71.91 & 72.39 \\
    & 1.0 & 72.12 & 72.46 \\
    \hline
    \multirow{6}{*}{\shortstack{ViT-S (DeiT)}}  
    & \textcolor{darkgray}{4.6} & \textcolor{darkgray}{79.82} & \textcolor{darkgray}{-} \\
    & 2.3 & 78.74 & 79.39 \\
    & 2.5 & 79.09 & 79.61 \\
    & 2.7 & 79.40 & 79.64 \\
    & 2.9 & 79.58 & 79.83 \\
    & 3.1 & 79.71 & 79.90 \\
    \hline
    \multirow{6}{*}{\shortstack{ViT-B (DeiT)}}  
    & \textcolor{darkgray}{17.6} & \textcolor{darkgray}{81.82} & \textcolor{darkgray}{-} \\
    & 8.7 & 78.98 & 80.61 \\
    & 10.0 & 80.63 & 81.17 \\
    & 10.4 & 80.97 & 81.30 \\
    & 11.5 & 81.50 & 81.59 \\
    & 12.5 & 81.75 & 81.80 \\
    \hline
    \end{tabular}
    \label{tab:full_results_deit}
\end{table}

\subsection{MAE Models}\label{sec:mae_models}
Also, Table~\ref{tab:full_results_mae} presents the complete outcomes of MAE~\cite{he2022masked}, encompassing both off-the-shelf models and fine-tuned models. We do not fine-tune ViT-H due to the constraints of computational resources.

\begin{table}[!th]
    \centering
    \caption{\textbf{Full MAE Results.} FT denotes fine-tuning the compressed model for 30 epochs. \textcolor{darkgray}{Gray} denotes the official un-compressed pre-trained models (Baseline).}
    \begin{tabular}{cccc}
    \hline
    \multirow{2}{*}{Model} & \multirow{2}{*}{FLOPs(G)} & \multicolumn{2}{c}{Acc.(\%)} \\
    & & w/o FT & w/ FT \\
    \hline
    \multirow{5}{*}{\shortstack{ViT-B (MAE)}}  
    & \textcolor{darkgray}{17.6} & \textcolor{darkgray}{83.72} & \textcolor{darkgray}{-} \\
    & 8.7 & 79.96 & 81.89 \\
    & 10.0 & 81.87 & 82.65 \\
    & 10.4 & 82.07 & 82.83 \\
    & 11.5 & 82.91 & 83.19 \\
    \hline
    \multirow{6}{*}{\shortstack{ViT-L (MAE)}}  
    & \textcolor{darkgray}{61.6} & \textcolor{darkgray}{85.95} & \textcolor{darkgray}{-} \\
    & 31.0 & 84.65 & 85.31 \\
    & 34.7 & 85.19 & 85.45 \\
    & 38.5 & 85.45 & 85.61 \\
    & 42.3 & 85.56 & 85.63 \\
    & 46.1 & 85.76 & 85.84 \\
    \hline
    \multirow{5}{*}{\shortstack{ViT-H (MAE)}}  
    & \textcolor{darkgray}{167.4} & \textcolor{darkgray}{86.88} & \textcolor{darkgray}{-} \\
    & 83.7 & 86.15 & - \\
    & 93.2 & 86.40 & - \\
    & 103.4 & 86.72 & - \\
    & 124.5 & 86.77 & - \\
   
    \hline
    \end{tabular}
    \label{tab:full_results_mae}
\end{table}

\begin{table*}[!th]
    \centering
    \caption{\textbf{Searched Compression Schedule.} We provide the block-wise kept tokens number for pruning and merging, respectively. }
    \setlength\tabcolsep{4pt}
    \begin{tabular}{ccc}
    \hline
    \multirow{2}{*}{Model} & \multirow{2}{*}{FLOPs(G)} & Compression Schedule  \\
    & & Prune \& Merge \\
    \hline
    \multirow{10}{*}{\shortstack{ViT-T (DeiT)}}  
    & \multirow{2}{*}{0.6} & [197,196,180,157,130,107,86,73,63,51,40,3] \\
    & & [197,187,167,139,114,90,76,66,57,45,37,3] \\
    \cline{2-3}
    & \multirow{2}{*}{0.7} & [197,196,194,180,150,123,98,82,68,60,52,3] \\
    & & [197,196,192,158,133,103,88,72,64,58,49,3] \\
    \cline{2-3}
    & \multirow{2}{*}{0.8} & [197,197,196,186,166,147,117,103,92,80,74,3] \\
    & & [197,196,190,172,154,125,107,96,84,78,70,3] \\
    \cline{2-3}
    & \multirow{2}{*}{0.9} & [197,197,196,190,182,166,141,125,113,105,99,3] \\
    & & [197,196,194,188,172,147,129,115,107,103,96,3] \\
    \cline{2-3}
    & \multirow{2}{*}{1.0} & [197,197,196,194,188,184,178,164,156,137,129,3] \\
    & & [197,197,196,190,186,154,148,156,145,135,125,3] \\
    \hline
    \multirow{10}{*}{\shortstack{ViT-S (DeiT)}}  
    & \multirow{2}{*}{2.3} & [197,192,168,143,121,105,92,74,62,45,33,3] \\
    & & [197,180,156,127,109,98,80,66,52,37,31,3] \\
    \cline{2-3}
    & \multirow{2}{*}{2.5} & [197,192,174,156,131,115,111,99,76,50,39,3] \\
    & & [197,186,160,133,119,107,96,82,66,43,37,3] \\
    \cline{2-3}
    & \multirow{2}{*}{2.7} & [197,196,182,160,141,127,115,105,88,66,49,3] \\
    & & [197,188,170,147,131,119,109,96,76,54,45,3] \\
    \cline{2-3}
    & \multirow{2}{*}{2.9} & [197,196,190,168,150,139,129,117,99,78,58,3] \\
    & & [197,194,176,156,141,133,121,107,88,64,56,3] \\
    \cline{2-3}
    & \multirow{2}{*}{3.1} & [197,197,194,180,160,147,137,129,115,92,76,3] \\
    & & [197,196,186,164,150,141,133,121,103,78,68,3] \\
    \hline
    \multirow{10}{*}{\shortstack{ViT-B (DeiT)}}  
    & \multirow{2}{*}{8.7} & [197,192,172,156,131,107,90,72,56,33,17,3] \\
    & & [197,178,162,141,115,96,78,60,47,21,13,3] \\
    \cline{2-3}
    & \multirow{2}{*}{10.0} & [197,194,182,168,148,129,113,101,82,49,25,3] \\
    & & [197,186,174,156,137,117,105,90,66,31,19,3] \\
    \cline{2-3}
    & \multirow{2}{*}{10.4} & [197,196,188,174,156,141,121,103,90,58,25,3]  \\
    & & [197,190,184,164,145,131,107,94,74,31,21,3]\\
    \cline{2-3}
    & \multirow{2}{*}{11.5} & [197,197,197,188,170,154,139,123,107,72,43,3] \\
    & & [197,197,192,178,158,145,127,111,94,50,35,3] \\
    \cline{2-3}
    & \multirow{2}{*}{12.5} & [197,197,196,192,184,170,154,139,129,94,58,3] \\
    & & [197,196,194,190,176,160,145,131,115,68,47,3] \\
    \hline
    \multirow{8}{*}{\shortstack{ViT-B (MAE)}}  
    & \multirow{2}{*}{8.7} & [197,192,166,143,119,103,86,72,54,43,37,23] \\
    & & [197,176,150,127,107,92,76,58,47,41,23,23] \\
    \cline{2-3}
    & \multirow{2}{*}{10.0} & [197,194,173,154,135,120,108,92,76,64,52,40] \\
    & & [197,181,160,141,122,116,95,80,67,60,40,40] \\
    \cline{2-3}
    & \multirow{2}{*}{10.4} & [197,195,179,159,140,122,105,100,89,71,56,45] \\
    & & [197,187,166,145,125,109,102,97,75,64,45,45]\\
    \cline{2-3}
    & \multirow{2}{*}{11.5} & [197,196,182,172,162,147,131,109,101,94,70,49] \\
    & & [197,192,178,164,150,133,115,101,96,82,49,49] \\
    \hline
    \end{tabular}
    \label{tab:searched_compression_schedule}
\end{table*}

\subsection{Seached Compression Schedule}\label{sec:searched_compression_schedule}
Table~\ref{tab:searched_compression_schedule} presents the detailed compression schedule that we searched. Notably, in contrast to DeiT models, MAE models tend to retain more tokens in the deep block. This is because DeiT classifies solely based on the class token, while MAE classifies based on the average of all image tokens.

\subsection{Compression Schedule Transfer}\label{sec:compression_rate_transfer}
\begin{figure}[!t]
    \centering
    \includegraphics[width=0.9\linewidth]{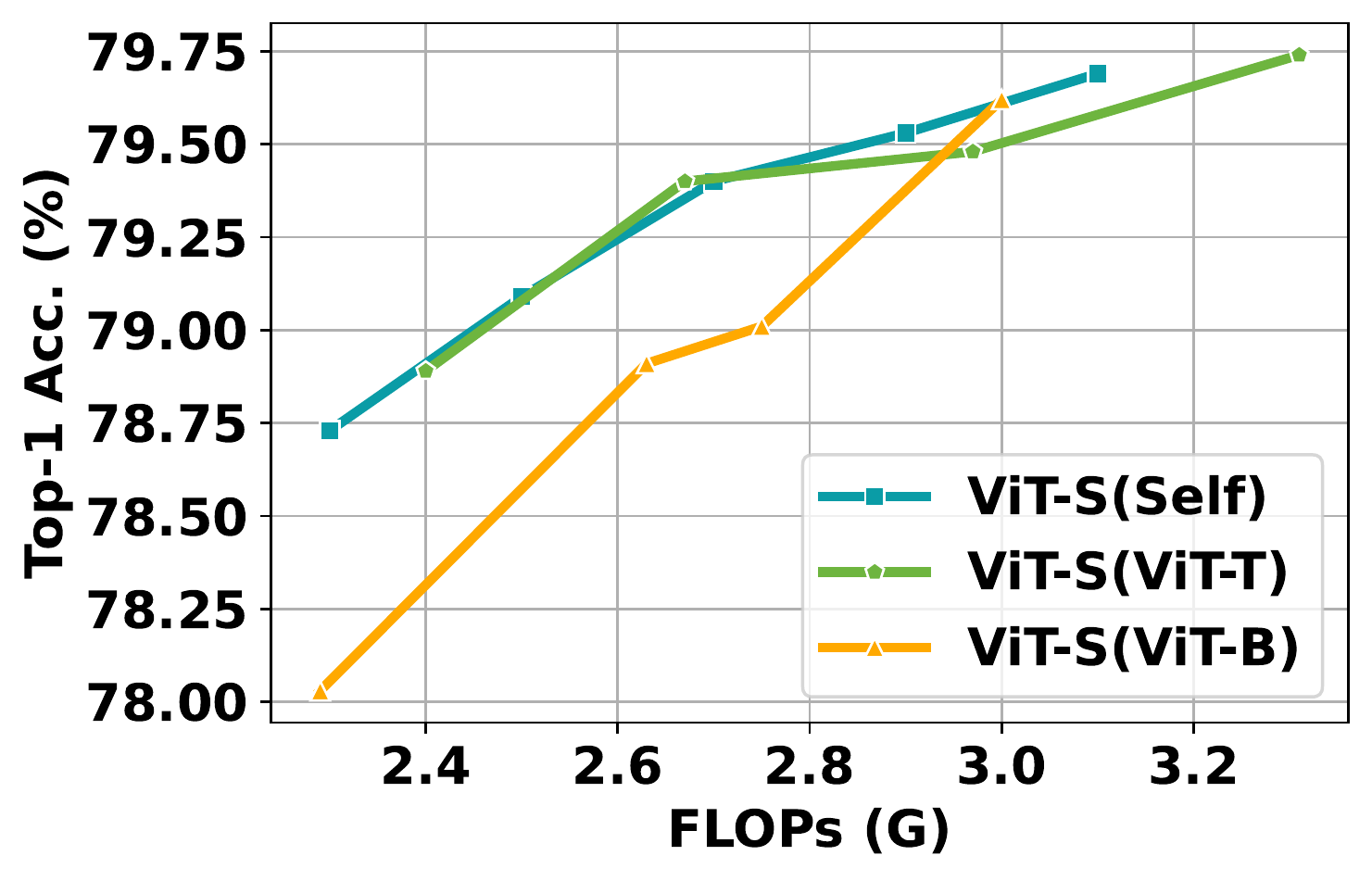}
    \caption{\textbf{Compression rate transfer}. Transferring the compression rate of ViT-B(DeiT) and ViT-T(DeiT) to ViT-S(DeiT). Self indicates the compression rate learn in ViT-S(DeiT).}
    \label{fig:compression_rate_transfer}
\end{figure}

The number of blocks in ViT-T, ViT-S, and ViT-B is 12, facilitating the transferability of their block-wise compression rates. To investigate this, we transferred the compression rates obtained from ViT-T and ViT-B to ViT-S, as illustrated in Fig.\,\ref{fig:compression_rate_transfer}. We can observe that the compression rate attained by ViT-S itself was optimal, whereas the transfer of compression rate from ViT-T to ViT-S resulted in a similar outcome. However, the transfer of compression rate from ViT-B to ViT-S leads to significantly poorer performance. This experiment verifies the ability of our proposed DiffRate to learn block-wise compression rates suitable for different network structures based on their features. Furthermore, it highlights that compression rates are somewhat transferable among similar network structures, such as transferring the compression rate from ViT-T to ViT-S.

\begin{table}[!th]
    \centering
    \caption{\textbf{Train from scratch.} Train compressed ViT-S (DeiT) from scratch with the official DeiT training receipt. }
    \setlength\tabcolsep{4pt}
    \begin{tabular}{ccccc}
    \hline
    Model & FLOPs & Throughput & Train Speed & Acc. \\
    \hline
    ViT-S (DeiT) & 4.6 & 5039 & 1$\times$ & 79.82 \\
    \rowcolor[gray]{0.9}DiffRate & 2.3 & 8901 & 1.8$\times$ & 79.41 \\
    \rowcolor[gray]{0.9}DiffRate & 2.9 & 6744 & 1.4$\times$ & 79.76 \\
    \hline
    \end{tabular}
    \label{tab:train_from_scratch}
\end{table}

\subsection{Train from Scratch}\label{sec:train_from_scratch}

The compressed model also has the capability to train from scratch using the searched compression rate. In this scenario, redundant tokens are directly eliminated, resulting in a faster training speed. As presented in Table\,\ref{tab:train_from_scratch}, DiffRate yields a 1.4$\times$ increase in training speed with only a $-0.06\%$ performance degradation.

\section{More Visualization}\label{sec:more_visualization}
\begin{figure*}[htbp]
    \centering
    \includegraphics[width=\linewidth]{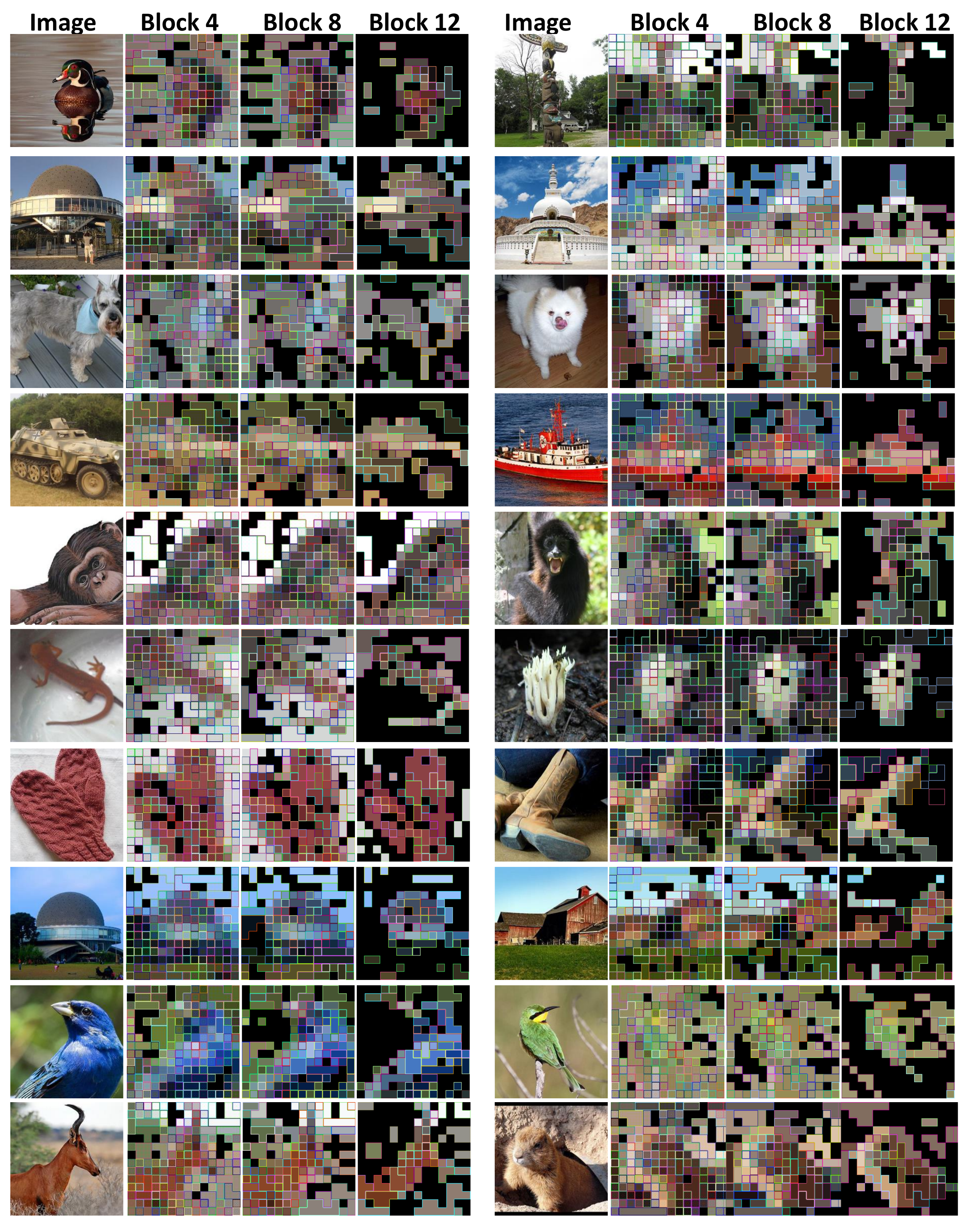}
    \caption{\textbf{More visualization.} Continuation of Fig\,\ref{fig:visualization}.}
    \label{fig:more_visualization}
\end{figure*}

We utilize the approach proposed by ToMe~\cite{bolya2022tome} to generate visualizations of the merging results. Specifically, we map each merged and pruned token back to its original input patch. To visualize merged tokens, we assign each input patch the average color of the merged tokens it belongs to and apply a random border color to distinguish tokens. For pruned tokens, we color their corresponding input patches black.
Fig.\,\ref{fig:more_visualization} provides additional examples of token compression on images, extending the visualizations shown in Fig.\,\textcolor{red}{5}. 
Our proposed DiffRate approach effectively identifies semantic objects, removes semantically irrelevant background tokens, and merges less-discriminative tokens in the foreground. By combining the advantages of pruning and merging through learnable compression rates, DiffRate can reduce the token count with minimal loss of information.

\section{Overhead of DiffRate}
In this section, we offer the analysis about computational overhead introduced by our proposed DiffRate framework. For a given number of input tokens, denoted by $N$, and a total of $L$ blocks, the reparameterization trick introduces $2NL$ parameters and $\frac{(N^2+5N)L}{2}$ FLOPs. For example, we consider DeiT-S with $N=196$ and $L=12$, which results in only $4.7$k additional parameters and $0.24$M additional FLOPs. 
Furthermore, we evaluated the FLOPs of the DiffRate module within one block, taking into account the number of input tokens ($N$), the embedding size ($C$), and the number of merging tokens ($N_{m}$). The FLOPs of the DiffRate module within a block can be approximated as $N\log{N} + (N-N_m+1)N_mC$. For instance, in the case of one DeiT-S block with $N=196$, $C=384$, and $N_m=20$, the FLOPs of the DiffRate module amount to $1.36$M. 
Overall, the overhead of DiffRate framework is negligible compared to the original $22$M parameters and $4.6$G FLOPs for DeiT-S.

\begin{figure}
    \centering
    \includegraphics{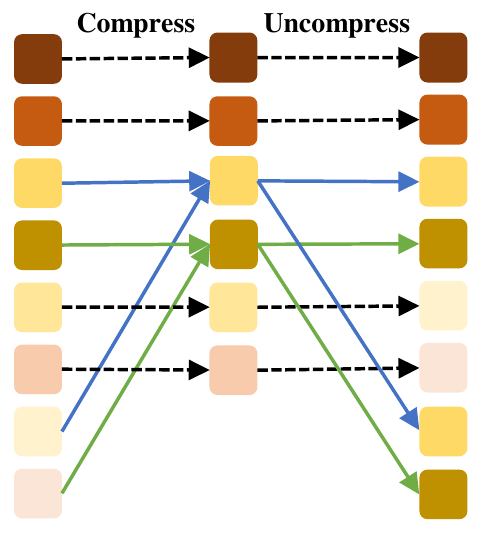}
    \caption{An example of token uncompression. We only consider token merging here.}
    \label{fig:uncompress}
\end{figure}

\begin{table}[!th]
    \centering
    \caption{Appling DiffRate to CAFormer-S36 on ImageNet-1k.}
    \begin{tabular}{cccc}
    \hline
    \multirow{2}{*}{Method} & \multirow{2}{*}{FLOPs(G)} & \multicolumn{2}{c}{Top-1 Acc.(\%)} \\
    & & w/o tuning & w/ tuning  \\
    \hline
    Baseline & 8.0 & 84.45 & - \\
    \hline
    \multirow{3}{*}{DiffRate} & 6.0 & 84.21 & 84.32\\
    & 5.6 & 83.92 & 84.21 \\
    & 5.2 & 83.49 & 84.03 \\
    \hline
    \end{tabular}
    \label{tab:hierarchical_experiments}
\end{table}

\begin{table}[!th]
    \centering
    \caption{Downstream semantic segmentation task using Semantic FPN with CAFormer-S36 backbone on ADE20K dataset. FLOPs is calculated under the input scale of $512 \times 704$.}
    \begin{tabular}{cccc}
    \hline
    Backbone & FLOPs (G) & mIoU (\%) & mAcc (\%) \\
    \hline
    CAFormer-S36 & 77.2 & 41.05 & 51.50 \\
    w/ DiffRate & 54.5 & 40.88 & 51.32 \\
    \hline
    \end{tabular}
    \label{tab:semantic}
\end{table}

\section{Extending to Hierarchical Architecture}\label{sec:hierachical}
In addition to the standard ViT~\cite{dosovitskiy2020image}, various ViT variants have emerged, including EfficientFormer~\cite{li2022efficientformer}, Swin Transformer~\cite{liu2021swin}, CAFormer~\cite{yu2022metaformer}, among others. These modern ViT variants commonly employ a hierarchical architecture, dividing the network into multiple stages and progressively reducing the resolution of the feature map.
Within hierarchical architectures, downsampling operations, such as convolution or pooling, are used to achieve the reduction in resolution. These operations rely on maintaining the spatial integrity of the feature map. However, traditional token compression operations disrupt the spatial integrity by reducing the number of tokens in an unstructured manner. As a result, conventional token compression techniques cannot be directly applied to hierarchical architectures.
To overcome this challenge, we propose a token uncompression module (see Fig.,\ref{fig:uncompress}) that restores the compressed token sequence by copying tokens based on their relationships, inspired by the approaches in ToMeSD~\cite{bolya2023tomesd} and TCFormer~\cite{zeng2022not}. By incorporating this module at the end of each stage, we can adapt our proposed DiffRate method to hierarchical architectures.
Specifically, we apply DiffRate to the CAFormer~\cite{yu2022metaformer}, a state-of-the-art ViT variant with four stages. DiffRate is applied only to the third stage, as the first two stages consist of convolution blocks, and the last stage incurs minimal computational cost (6\% in CAFormer-S36).
As shown in Table\,\ref{tab:hierarchical_experiments}, DiffRate achieves a significant 25\% reduction in FLOPs with a marginal sacrifice of 0.24\% in accuracy without additional fine-tuning. Moreover, when fine-tuned, DiffRate achieves a remarkable 35\% reduction in FLOPs with a minimal accuracy drop of only 0.42\%.

\section{Extending to Downstream Tasks}
By incorporating the introduced uncompression module in Fig.\,\ref{fig:uncompress}, DiffRate can also be applied to the model for downstream tasks. We begin by transferring the uncompressed pre-trained CAFormer-S36 model to ADE20K~\cite{zhou2017scene}, following the settings of PVT~\cite{wang2021pyramid}. Subsequently, DiffRate is applied to the third stage of the trained segmentation model without further fine-tuning.
The results in Table,\ref{tab:semantic} demonstrate that DiffRate achieves a significant 30\% reduction in FLOPs with negligible performance degradation.

\end{document}